\documentclass[twoside]{article}

\usepackage[accepted]{aistats2025}

\usepackage[utf8]{inputenc} 
\usepackage[T1]{fontenc}    
\usepackage{hyperref}       
\usepackage{url}            
\usepackage{booktabs}       
\usepackage{amsfonts}       
\usepackage{nicefrac}       
\usepackage{microtype}      
\usepackage{xcolor}         

\usepackage[utf8]{inputenc} 
\usepackage[T1]{fontenc}    
\usepackage{hyperref}
\usepackage{xcolor}
\hypersetup{
    colorlinks,
    linkcolor={red!50!black},
    citecolor={magenta},
    urlcolor={blue!90!black}
}

\usepackage{url}            
\usepackage{booktabs}       
\usepackage{amsfonts}       
\usepackage{nicefrac}       
\usepackage{microtype}      
\usepackage{xcolor}         

\usepackage{graphicx}
\usepackage{amsmath}
\usepackage{amssymb}
\usepackage{booktabs}
\usepackage{amsthm}

\usepackage{lipsum}  

\newcommand{\RR}{\mathbb{R}} 
\newtheorem{theorem}{Theorem}
\newtheorem{proposition}{Proposition}
\newtheorem{lemma}{Lemma}

\usepackage[capitalize]{cleveref}
\crefname{section}{Sec.}{Secs.}
\Crefname{section}{Section}{Sections}
\Crefname{table}{Table}{Tables}
\crefname{table}{Tab.}{Tabs.}

\usepackage{hyperref}       
\usepackage{url}            
\usepackage{booktabs}       
\usepackage{multirow}

\usepackage{amsfonts}       
\usepackage{nicefrac}       
\usepackage{microtype}      

\usepackage{amsmath}
\usepackage{amssymb}
\usepackage{mathtools}
\usepackage{amsthm}

\usepackage{enumitem}

\usepackage{amssymb}

\usepackage{dsfont}

\usepackage{overpic}
\usepackage{subcaption}
\usepackage{empheq}

\newcommand{\bW}{{\bf W}}
\newcommand{\bU}{{\bf U}}

\newcommand{\ba}{{\bf a}}
\newcommand{\bg}{{\bf g}}
\newcommand{\bh}{{\bf h}}

\newcommand{\bu}{{\bf u}}

\newcommand{\bx}{{\bf x}}
\newcommand{\by}{{\bf y}}
\newcommand{\bz}{{\bf z}}

\definecolor{olive}{rgb}{0.6, 0.6, 0.2}
\definecolor{sand}{rgb}{0.8666666666666667, 0.8, 0.4666666666666667}
\definecolor{wine}{rgb}{0.5333333333333333, 0.13333333333333333, 0.3333333333333333}
\definecolor{deblue}{RGB}{11,132,147}
\definecolor{ocra}{RGB}{204, 119, 34}

\definecolor{darkcyan}{RGB}{0,139,139}

\usepackage[most]{tcolorbox}

\newtcolorbox{CatchyBox}[2][]{
	lower separated=false,
	colback=white!80!sand!90!ocra,
	colframe=white, fonttitle=\bfseries,
	colbacktitle=white!50!sand!90!ocra,
	coltitle=black,
	enhanced,
	attach boxed title to top left={xshift=.02\linewidth,yshift=-4mm},
	title=#2,#1}

\usepackage{wrapfig}

\usepackage[authoryear]{natbib}

\begin{document}

\runningtitle{Gated RNNs with Weighted Time-Delay Feedback}

\twocolumn[

\aistatstitle{Gated Recurrent Neural Networks \\ with Weighted Time-Delay Feedback}

\aistatsauthor{N. Benjamin Erichson \And Soon Hoe Lim \And  Michael Mahoney }

\aistatsaddress{
ICSI and LBNL \\  \And KTH and Nordita \And UC Berkeley, ICSI and LBNL} ]

\begin{abstract}
  In this paper, we present a novel approach to modeling long-term dependencies in sequential data by introducing a gated recurrent unit (GRU) with a weighted time-delay feedback mechanism. Our proposed model, named $\tau$-GRU, is a discretized version of a continuous-time formulation of a recurrent unit, where the dynamics are governed by delay differential equations (DDEs). We prove the existence and uniqueness of solutions for the continuous-time model and show that the proposed feedback mechanism can significantly improve the modeling of long-term dependencies. Our empirical results indicate that $\tau$-GRU outperforms state-of-the-art recurrent units and gated recurrent architectures on a range of tasks, achieving faster convergence and better generalization.
\end{abstract}

\section{INTRODUCTION}

Recurrent neural networks (RNNs) and their variants are flexible gradient-based methods specially designed to model sequential data. 
Models of this type can be viewed as dynamical systems whose temporal evolution is governed by a system of differential equations driven by an external input.
In fact, there is a long tradition of formulating continuous-time variants of RNN~\citep{pineda1988dynamics}. 
In this setting, the data are formulated in continuous-time, i.e., inputs are defined by the function $\bx = \bx(t) \in \mathbb{R}^p$ and the targets are defined as $\by = \by(t) \in \mathbb{R}^q$, where $t$  denotes continuous time.
In this way, one can, for instance, employ a nonautonomous ordinary differential equation (ODE) to model the dynamics of the hidden states $\bh(t)\in\mathbb{R}^d$: 
\begin{equation*}
     \frac{d \bh(t)}{d t}\,\,\, = \,\,\, f(\bh(t),\, \bx(t);\, \boldsymbol{\theta}).
\end{equation*}
Here, $f: \mathbb{R}^d\times  \mathbb{R}^p \rightarrow  \mathbb{R}^d$ is a function that is parameterized by a neural network (NN) with learnable weights $\boldsymbol{\theta}$. 
A prototypical choice for $f$ is the $\tanh$ recurrent unit:
\begin{equation*}
    f(\bh(t),\, \bx(t); \boldsymbol{\theta}) := \tanh(\bW \bh(t) + \bU \bx(t) + {\bf b}),
\end{equation*}
where $\bW \in \mathbb{R}^{d\times d}$ denotes a hidden-to-hidden weight matrix, $\bU\in \mathbb{R}^{d \times p}$ an input-to-hidden weight matrix, and ${\bf b}$ a bias term.
With this continuous-time formulation in hand, one can then use tools from dynamical systems theory to study the dynamical behavior of the model and to motivate mechanisms that can prevent rapidly diverging or converging dynamics. 
For instance,~\citet{chang2018antisymmetricrnn}  proposed a parametrization of the hidden matrix as an antisymmetric matrix to ensure stable hidden state dynamics, and~\citet{erichson2020lipschitz} relaxed this idea to improve the model expressivity. 
More recently,~\cite{rusch2022long} has proposed an RNN architecture based on a suitable time-discretization of a set of coupled multiscale ODEs.

In this work, we consider the use of input-driven nonlinear delay differential equations (DDEs) to model the dynamics of hidden states:
\begin{align*}
    \frac{d \bh(t)}{d t} \,\,\, = \,\,\, f(\bh(t),\, \bh(t-\tau),\, \bx(t);\, \boldsymbol{\theta}),
\end{align*}
where $\tau$ is a constant that indicates the delay (i.e., time-lag). Here, the time derivative is described by a function $f: \mathbb{R}^d\times \mathbb{R}^d\times  \mathbb{R}^p \rightarrow  \mathbb{R}^d$ that explicitly depends on states from the past. 
Previous work~\citep{lin1996learning} has shown that delay units can improve performance in long-term dependency problems \citep{pascanu2013difficulty}, i.e., problems for which the desired model output depends on inputs presented at times far in the past.

\begin{figure*}[!t] 
	\begin{CatchyBox}{$\tau$-GRU}\vspace{+0.3cm}
		\textbf{Continuous-time formulation of $\tau$-GRU:}\\
		\begin{minipage}[h]{0.95\linewidth}
			\begin{align}
			\label{eq:tdRNN_de}
			\frac{d \, \bh(t)}{dt}  
			= \underbrace{g(\bh(t), \bx(t))}_{\text{gating}} \odot \left( \underbrace{u(\bh(t),\, \bx(t))}_\text{instantaneous dynamics} \, + \,\,\, \underbrace{\textcolor{wine}{a(\bh(t), \bx(t))}  \odot \textcolor{darkcyan}{z(\bh(t-\tau),\, \bx(t))}}_\text{weighted time-delayed feedback} \,\,\, - \,\,\, \bh(t) \right) 
			\end{align}
			
		\end{minipage}\vspace{+0.3cm}
		
		\textbf{Discrete-time formulation of $\tau$-GRU:}\\
		\begin{minipage}[h]{0.57\linewidth}
			\begin{equation}    
			\bh_{n+1} = (1-\bg_n) \odot \bh_n + \bg_n \odot (\bu_n +  \textcolor{wine}{\ba_n} \odot \textcolor{darkcyan}{\bz_n}) \label{eq:ourdRNN}  
			\end{equation}
			with
			\vspace{-0.3cm}
			\begin{align}
			\bu_n & = u(\bh_n, \bx_n) := \text{tanh}(\bW_1 \bh_n + \bU_1 \bx_n) \label{eq:3} \\
			\textcolor{darkcyan}{\bz_n} & = \textcolor{darkcyan}{z(\bh_{l}, \bx_n)} := \text{tanh}(\bW_2 \bh_{l} + \bU_2 \bx_{n}) \\
			\bg_n & = g(\bh_n, \bx_n) :=   \text{sigmoid}(\bW_3 \bh_n + \bU_3 \bx_n) \\
			\textcolor{wine}{\ba_n} & = \textcolor{wine}{a(\bh_n, \bx_n)} :=\text{sigmoid}(\bW_4 \bh_{n} + \bU_4 \bx_{n}) \label{eq:6}
			\end{align}
			\hfill
		\end{minipage}
		\hfill
		\begin{minipage}[h]{.49\linewidth}\small
			\centering
			\begin{tabular}{l|c|c}
				input & $\bx$ & $\mathbb{R}^{p}$\\\hline
				time index & $t$ & $\mathbb{R}$ 	 \\\hline
				time delay & $\tau$ & $\mathbb{R}$ 	 \\\hline		
				hidden state & $\bh$ & $\mathbb{R}^{d}$\\\hline
				hidden-to-hidden matrix & $\bW_i$ & $\mathbb{R}^{d \times d}$\\\hline
				input-to-hidden matrix & $\bU_i$ & $\mathbb{R}^{d\times p}$\\\hline	
				decoder matrix & ${\bf V}$ & $\mathbb{R}^{q\times d}$\\\hline
			\end{tabular} \\
			\vspace{0.2cm}
			$\bh_n \approx \bh(t_n)$, $t_n = n \Delta t$, $n=0,1,\dots$ \\
			\vspace{0.1cm}
			$l := n - \lfloor \tau/\Delta t \rfloor$
		\end{minipage}	
	\end{CatchyBox}
\end{figure*}

In more detail, we propose a novel continuous-time nonlinear recurrent unit, given in Eq.~\eqref{eq:tdRNN_de}, that is composed of two parts:
(i) a component $u(\bh(t),\bx(t))$ that explicitly models instantaneous dynamics; and (ii) a component $z(\bh(t-\tau), \bx(t))$ that provides time-delayed feedback to account for noninstantaneous dynamics. 
Feedback also helps to propagate gradient information more efficiently, thus reducing the issue of vanishing gradients.
In addition, we introduce $a(\bh(t),\bx(t))$ to weight the importance of the feedback component-wise, which helps to better model different time scales. 
By considering a suitable time-discretization scheme of this continuous-time setup, we obtain a gated recurrent unit (GRU), given in Eq. \eqref{eq:ourdRNN}, which we call $\tau$-GRU. The individual parts are described by Eq.~\eqref{eq:3}-\eqref{eq:6}, where $\bg_n$ and $\ba_n$ resemble commonly used gating functions.

Although nonlinear RNNs have been widely used to model sequences for decades, other models have also been developed more recently. However, there are trade-offs that different sequence models have. Our proposed RNN model works better than Transformers \citep{vaswani2017attention, tay2022efficient} and state space models (SSMs) \citep{gu2021combining, gu2020hippo, gu2021efficiently} in the small data regime while being parameter efficient (see also App. 
\ref{sect_AppH} for more details) -- {\it we emphasize that this is the regime on which we focus in the present work}. The computational cost of the model scales linearly with the sequence length and is typically faster than the Transformers at inference time, even for modest sequence lengths. However, it is still slow to optimize due to the inherently sequential nature of the computation and is therefore hard to scale. Transformers and SSMs are easier to scale and excel in the big data regime but typically need a much larger number of trainable parameters and tend to overfit in the small data regime. While Transformers do not have to face the sequential training issue of recurrent models (and can be trained in parallel), the computational and memory cost of Transformers scales quadratically with the sequence length, and can therefore be very expensive to deploy on long sequences.

\textbf{Main Contributions.}
Our key contributions are as follows. 

\begin{figure}[!t]
	\centering	\includegraphics[width=0.45\textwidth]{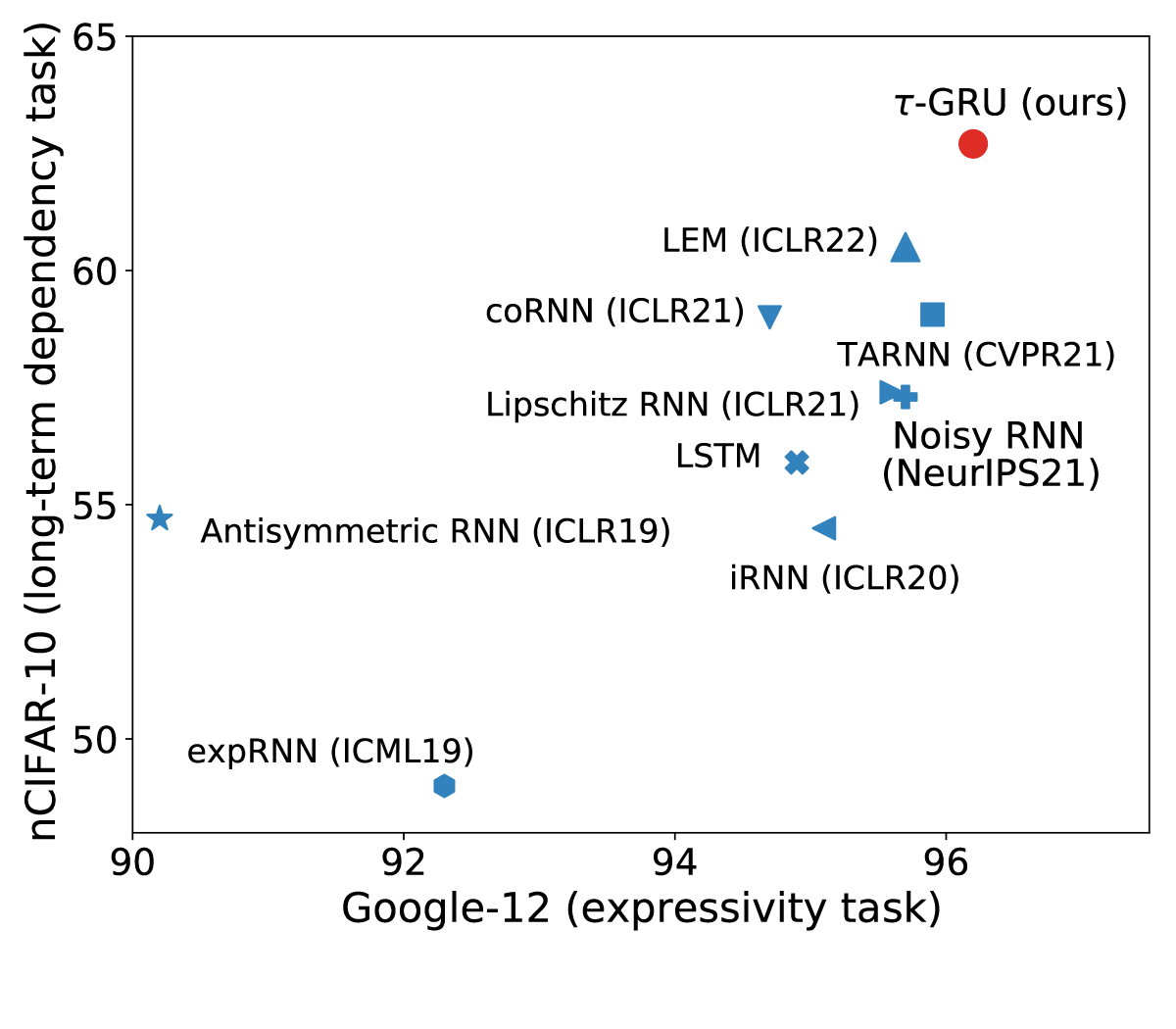}\vspace{-0.6cm}
	\caption{Test accuracy for nCIFAR~\citep{chang2018antisymmetricrnn} versus Google-12~\citep{warden2018speech}.  nCIFAR requires a recurrent unit with long-term dependency capabilities, while Google-12 requires a highly expressive unit.  Our  $\tau$-GRU is able to improve performance on both tasks, relative to existing state-of-the-art alternatives, including LEM~\citep{rusch2022long}.}
	\label{fig:intro_fig}
\end{figure}

\vspace{-0.3cm}
\begin{itemize}[leftmargin=*]
\item{\bf Design:} 
We introduce a novel gated recurrent unit, which we call $\tau$-GRU, which incorporates a weighted time-delay feedback mechanism to reduce the problem of vanishing gradients. This model is motivated by nonlinear DDEs, and it is obtained by discretizing the continuous Eq. \eqref{eq:tdRNN_de}.

\vspace{-0.1cm}
\item {\bf Theory:} 
We show that the continuous-time $\tau$-GRU model has a unique well-defined solution (see Theorem~\ref{thm_exist2main}).
Moreover, we provide intuition and analysis to understand how the introduction of delays in $\tau$-GRU can act as a buffer to help alleviate the problem of vanishing gradients, thus improving the ability to retain information long in the past.
See Proposition \ref{prop_delaymain} for a simplified setting and Proposition \ref{app_prop} in App. \ref{app:gradbound} for  $\tau$-GRU.

\vspace{-0.1cm}
\item {\bf Experiments:} 
We provide empirical results to demonstrate the performance of $\tau$-GRU on a variety of benchmark tasks. 
We show that $\tau$-GRU converges faster during training and can achieve improved generalization performance. Moreover, we demonstrate that it provides favorable trade-offs between effectiveness in dealing with long-term dependencies and expressivity in the considered tasks. See Figure \ref{fig:intro_fig} for an illustration of this. 
\end{itemize}

Our main focus here is on {\it shallow (single-layer) nonlinear RNNs}, in particular, we provide a theoretically supported approach to improve upon existing RNN models, not so much about achieving state-of-the-art results among all the possible sequence models out there.

\section{RELATED WORK}

In this section, we discuss recent RNN advances that have been shown to outperform classic architectures such as Long Short Term Memory (LSTM) networks~\citep{hochreiter1997long} and the GRUs~\citep{cho2014properties}. We also briefly discuss previous work on incorporating delays into NNs. 

\noindent 
\textbf{Unitary and orthogonal RNNs.} \
The seminal work \citep{arjovsky2016unitary} introduced a recurrent unit in which the hidden matrix is constructed as the product of unitary matrices.
This enforces that the eigenvalues lie on the unit circle.
This, in turn, prevents vanishing and exploding gradients, thereby enabling the learning of long-term dependencies. 
However, such unitary RNNs suffer from limited expressivity, since the construction of the hidden matrix is restrictive~\citep{azencot2021differential}. 
Work by~\cite{wisdom2016full} and~\cite{vorontsov2017orthogonality} partially addressed this issue by considering the Cayley transform on skew-symmetric matrices; and 
work by~\cite{lezcano2019cheap,lezcano2019trivializations} leveraged skew-Hermitian matrices to parameterize the orthogonal group to improve expressiveness. 
The expressiveness of RNNs has been further improved by considering nonnormal hidden matrices~\citep{kerg2019non}.

\noindent 
\textbf{Continuous-time nonlinear RNNs.} \ 
The work on Neural ODEs~\citep{chen2018neural} and variants~\citep{kidger2020neural,queiruga2021stateful,xia2021heavy,hasani2022closed} have motivated the formulation of several modern continuous-time RNNs, which are expressive and have good long-term memory. 
The work by~\cite{chang2018antisymmetricrnn} used an antisymmetric matrix to parameterize the hidden-to-hidden matrix in order to obtain stable dynamics. 
In~\citep{Kag2020RNNs}, a modified differential equation was considered, which allows one to update the hidden states based on the difference between the predicted and previous states. 

\cite{rusch2021coupled} demonstrated that long-term memory can be improved by modeling hidden dynamics by a second-order system of ODEs, which models a coupled network of controlled forced and damped nonlinear oscillators. Another approach to improve long-term memory was motivated by a time-adaptive discretization of an ODE~\citep{kag2021time}.
The expressiveness of continuous-time RNNs has been further improved by introducing a suitable time-discretization of a set of multiscale ODEs~\citep{rusch2022long}.
Lastly, \cite{lim2021noisy} studied noise-injected RNNs that can be viewed as discretizations of stochastic differential equations driven by input data. In this case, the noise can help stabilize the hidden dynamics during training and improve robustness. 

\noindent 
\textbf{Using delays in NNs.}
The idea of introducing delays into NNs goes back to  \citep{waibel1989phoneme, lang1990time}. 
Several works followed: \cite{kim1998time} considered a time-delayed RNN model that is suitable for temporal correlations and prediction of chaotic and financial time series; and delays were also incorporated into the nonlinear autoregressive  with exogenenous inputs (NARX) RNNs \citep{lin1996learning}. 
More recently, \cite{zhu2021neural} introduced delay terms in Neural ODEs and demonstrated their approximation capabilities. 
In particular, this model can learn delayed dynamics where the trajectories in the lower-dimensional phase space could be mutually intersected, while the standard Neural ODEs~\citep{chen2018neural} are not able to do so.

\noindent  \textbf{State-space models.} Recently, state-space models (SSMs)~\citep{gu2021efficiently, gu2021combining,yu2024tuning} have emerged as alternatives to RNNs due to their strong performance on long-sequence tasks. SSMs have demonstrated state-of-the-art results in diverse applications in video, audio, and time-series processing, often exceeding traditional LSTM and Transformer architectures while offering significant speed and memory efficiency~\citep{gu2023mamba}. 

In contrast to RNNs, which rely on sequential processing and suffer from slow training, SSMs exploit linear time-invariant (LTI) dynamics, enabling parallel computation in the time domain~\citep{smith2022simplified} or frequency domain~\citep{yu2023robustifying,yu2024hope}. Such parallelization significantly improves computational efficiency and scalability for long sequences.

\section{METHOD}

In this section, we provide an introduction to DDEs; then, we motivate the formulation of our DDE-based models, in continuous and discrete time; and finally, we propose a weighted time-delay feedback architecture. 

\textbf{Notation.} $\odot$ denotes the Hadamard product, $| v |$ denotes the vector norm for the vector $v$, $\| A \|$ denotes operator norm for the matrix $A$, $\sigma$ and $\hat{\sigma}$ (or sigmoid) denote the function $\tanh$ and sigmoid, respectively; and $\lceil x \rceil$ and $\lfloor x \rfloor$ denote the ceiling and floor function in $x$.

\subsection{Delay Differential Equations}

DDEs are an important class of dynamical systems that arise in natural and engineering sciences~\citep{smith2011introduction, erneux2009applied, keane2017climate}. In these systems, a feedback term is introduced to adjust the system non-instantaneously, resulting in delays in time. In mathematical terms, the derivative of the system state depends explicitly on the past value of the state variable.
Here, we focus on DDEs with a single discrete delay 
\begin{equation} \label{eq_gen_dde}
\dot{h} = F(t, h(t), h(t-\tau)),
\end{equation}
with $\tau > 0$, where $F$ is a continuous function. Due to the presence of the delay term, we need to {\it specify an initial
function} that describes the behavior of the system prior to the initial time $t=0$. For DDE, it would be a function $\phi$ defined in $[-\tau, 0]$.  Hence, a DDE numerical solver must save all the information needed to approximate delayed~terms.

Instead of thinking the  solution of the DDE as consisting of a sequence of values of $h$ at increasing values of $t$, as one would do for ODEs, it is more fruitful to view it as a mapping of functions on the interval $[t -\tau,t]$ into functions on the interval $[t,t + \tau]$, i.e., as a sequence of functions  defined over a set of contiguous time intervals of length $\tau$. 
Since the state of the system at time $t \geq  0$ must contain all the information necessary to determine the solution for future times $s \geq  t$, it should  contain the initial condition $\phi$.

More precisely, the DDE is a functional differential equation with the state space $C := C([-\tau, 0], \mathbb{R}^d)$. This state space is the Banach space of continuous functions from $[-\tau, 0]$ into $\mathbb{R}^d$, with the topology of uniform convergence. It is equipped with the norm $\|\phi \| := \sup \{ |\phi(\theta)| : \theta \in [-\tau, 0] \}$. 
In contrast to the ODEs (with $\tau = 0)$ whose state space is finite-dimensional, DDEs are generally infinite-dimensional dynamical systems. 
Various aspects of DDEs have been studied, including their solution properties \citep{hale2013introduction, asl2003analysis}, dynamics \citep{lepri1994high, baldi1994delays} and stability \citep{marcus1989stability, belair1993stability, liao2002delay, yang2014exponential, park2019dynamic}. 

\subsection{Continuous-Time \texorpdfstring{$\tau$}{tau}-GRUs}

The basic form of a time-delayed RNN is 
\begin{align}\label{eq:simple_model}
\begin{split}
    \dot{\bh} =\sigma(\bW_1 \bh(t) + \bW_2 \bh(t-\tau)+ \bU \bx(t) ) - \bh(t),
\end{split}
\end{align}
for $t \geq 0$,  and $\bh(t) = 0$ for $t \in [-\tau, 0]$, with the output $\by(t) = {\bf V} \bh(t)$. In this expression, $\bh \in \RR^d$ denotes the hidden states, $f: \RR^d \times \RR^d \times \RR^{p} \to \RR^d$ is a nonlinear function, and $\sigma: \RR \to (-1,1)$ denotes the tanh  activation function applied component-wise. The matrices $\bW_1, \bW_2 \in \RR^{d \times d}$, $\bU \in \RR^{d \times p}$ and ${\bf V} \in \RR^{q \times d}$ are learnable parameters, and $\tau \geq 0$ denotes the discrete time-lag. 
For notational brevity, we omit the bias term here, assuming that it is included in $\bW_1$.

It is important to equip RNNs with the mechanism to better represent a large number of scales, as discussed by \cite{tallec2018can} and more recently by~\cite{rusch2022long}. 
Therefore, we follow \citep{tallec2018can} and consider a time warping function $c:\mathbb{R}^d\rightarrow \mathbb{R}^d$ which we define to be a parametric function  $c(t)$. Using the reasoning in~\citep{tallec2018can}, and denoting $t_\tau := t-\tau$,  we can formulate the following continuous-time delay recurrent unit
\begin{align*}\label{eq:simple_model}
    \dot{\bh} = \frac{d c(t)}{d t}\left[ \sigma(\bW_1 \bh(t) + \bW_2 \bh(t_\tau) + \bU_1 \bx(t)) - \bh(t)\right].
\end{align*}

Now, we need a learnable function to model $\frac{d c(t)}{d t}$. A natural choice is to consider a standard gating function, which is a universal approximator, taking the form
\begin{equation}
    \frac{d c(t)}{d t} = \hat{\sigma}(\bW_3 \bh_t + \bU_3 \bx_t) =: \bg(t), 
\end{equation}
where $\bW_3 \in \RR^{d \times d}$ and  $\bU_3 \in \RR^{d \times p}$ are learnable parameters, and  $\hat{\sigma}: \RR \to (0,1)$ is the component-wise sigmoid function.

\subsection{Discrete-Time \texorpdfstring{$\tau$}{tau}-GRUs}

To learn the weights of the recurrent unit, a numerical integration scheme can be used to discretize the continuous model. Specifically, we discretize the time as $t_n = n \Delta t$ for $n = -\lfloor \tau/\Delta t \rfloor, \dots, -1, 0, 1, \dots$, and approximate the solution $(\bh(t))$ to Eq. \eqref{eq:simple_model} by the sequence $(\bh_n = \bh(t_n))$, given by $\bh_n = 0$ for $n = -\lfloor \tau/\Delta t \rfloor, \dots, 0$, and
\begin{align}
    \bh_{n+1} &= \bh_n + \int_{t_n}^{t_n+\Delta t} f(\bh(s),\, \bh(s-\tau), \bx(s)) \,\mathrm{d}s  \\
    &\approx \bh_n + \mathtt{scheme}[f,\,\bh_n,\,\bh_{l}, \Delta t],
\end{align}
for $n=0,1,\dots$, where the subscript $n$ denotes discrete time indices, $l := n - \lfloor \tau/\Delta t \rfloor$,  and $\Delta t$ represents the time difference between a pair of consecutive elements in the input sequence. 
In addition, $\mathtt{scheme}$ refers to a numerical integration scheme whose application yields an approximate solution for the integral.
Using the forward Euler scheme and choosing $\Delta t = 1$ gives:
\begin{equation} \label{eq_simpleDRNN}
   \bh_{n+1} = (1-\bg_n) \odot \bh_n + \bg_n \odot \sigma(\bW_1 \bh_n + \bW_2 \bh_{l} + \bU \bx_n).
\end{equation}
Note that this discretization corresponds to the leaky integrator described by~\cite{jaeger2007optimization} and we can, in principle, use other values of $\Delta t$.  We choose forward Euler because it is computationally efficient. Higher-order integrator schemes require more function evaluations, while tending not to help improve performance \citep{queiruga2020continuous}. 

It can be shown that \eqref{eq_simpleDRNN} is a universal approximator of a large class of open dynamical systems with delay (see Theorem \ref{thm_uat}).
However, the performance of this architecture cannot outperform existing RNN architectures on a number of tasks. To improve the performance, we next propose a modified model.

\subsection {Discrete-Time \texorpdfstring{$\tau$}{tau}-GRUs with a Weighted Time-Delay Feedback Architecture}

In this work, we propose to model the hidden dynamics using a mixture of a standard recurrent unit and a delay recurrent unit. To this end, we replace the $\sigma$ in Eq. \eqref{eq_simpleDRNN} by
\begin{equation} \label{eq_motivate}
    \bu_n + \ba_n \odot \bz_n,
\end{equation}
so that we yield a new GRU that takes the form
\begin{equation}
   \bh_{n+1} = (1-\bg_n) \odot \bh_n + \bg_n \odot \left( \bu_n + \ba_n \odot \bz_n \right).
\end{equation}
Here, $\bu_n$ describes the standard recurrent unit $\bu_n=\text{tanh}(\bW_1 \bh_n + \bU_1 \bx_n),$ and $\bz_n$ describes the delay recurrent unit $\bz_n  = \text{tanh}(\bW_2 \bh_{l} + \bU_2 \bx_{n}).$
Further, the gate $\bg_n$ is a learnable vector-valued coefficient 
$\bg_n  =\text{sigmoid}(\bW_3 \bh_{n} + \bU_3 \bx_{n}).$
The weighting term $\ba_n$ is also a vector-valued coefficient 
$\ba_n  =\text{sigmoid}(\bW_4 \bh_{n} + \bU_4 \bx_{n}),$
with the learnable parameters $\bW_4 \in \RR^{d \times d}$ and $\bU_4  \in \RR^{d \times p}$, that weights the importance of the time-delay feedback component-wise for the task on hand. 

From the design point of view, Eq. \eqref{eq_motivate} can be motivated by the sigmoidal coupling used in Hodgkin-Huxley type neural models (see Eq. (1)-(2) and Eq. (4) in \citep{campbell2007time}). Importantly, Eq. \eqref{eq_motivate} provides a flexible mechanism to allow the network to model a mixture of instantaneous term and a delay term for the non-linearity, thereby increasing the expressivity of the model. Note that Eq. \eqref{eq_simpleDRNN} also works well, but could not outperform other RNN models due to fewer trainable parameters. Our empirical evaluations on a wide range of experiments confirm the advantages of having this mechanism.

\section{THEORY}

In this section, we define the notion of a solution for DDEs and show that continuous-time $\tau$ -GRU has a unique solution. Moreover, we provide intuition and analysis to understand how the delay mechanism can help improve modeling long-term dependencies. 

\subsection{Existence and Uniqueness of Solution for Continuous-Time \texorpdfstring{$\tau$}{tau}-GRUs}

Since we must know $h(t + \theta)$, $\theta \in [-\tau, 0]$ in order to determine the solution of the DDE \eqref{eq_gen_dde} for $s > t$, we call the state of the dynamical system at time $t$ the element of $C$ which we denote as $h_t$,  defined as
$h_t(\theta) := h(t+\theta)$ for $\theta \in [-\tau, 0]$. 
The trajectory of the solution can thus be viewed as the curve $t \to h_t$ in the state space $C$. In general, DDEs can be formulated as the following initial value problem (IVP) for the nonautonomous system \citep{hale2013introduction}:
\begin{equation}  \label{eq_genddemain}
    \dot{h}(t) = f(t, h_t), \ t \geq t_0,
\end{equation}
where $h_{t_0} = \phi \in C$ for some initial time $t_0 \in \mathbb{R}$, and $f: \mathbb{R} \times C \to \mathbb{R}^d$ is a given continuous function. The above equation describes a general type of system, including ODEs ($\tau = 0$) and DDEs of the form $\dot{h}(t) = g(t, h(t), h(t-\tau))$ for some continuous function $g$. 

We say that a function $h$ is a solution of Eq. (\ref{eq_genddemain}) on $[t_0 - \tau, t_0 + A]$ if there exist $t_0 \in \mathbb{R}$ and $A > 0$ such that $h \in C([t_0 - \tau, t_0 + A), \mathbb{R}^d)$, $(t, h_t) \in \mathbb{R} \times C$ and $h(t)$
satisfies Eq. (\ref{eq_genddemain}) for $t \in [t_0, t_0 + A)$. It can be shown that (see, for instance, Lemma 1.1 in \citep{hale2013introduction}) if $f(t,\phi)$ is continuous, then finding a solution of Eq. (\ref{eq_genddemain}) is equivalent to solving the integral equation: $h_{t_0} = \phi$, 
\begin{equation}
    h(t) = \phi(0) + \int_{t_0}^t f(s, h_s) \  \rm{d}s, \ t \geq t_0.
\end{equation}

We now provide the existence and uniqueness results for the continuous-time $\tau$-GRU model, assuming that the input $x$ is continuous in $t$. Defining the state $h_t \in C$ as $h_t(\theta) := h(t+\theta)$ for $\theta \in [-\tau, 0]$ as before,  the DDE describing the $\tau$-GRU model can be formulated as the following IVP: 
\begin{equation} \label{eq_gendde} 
    \dot{h} = - h(t) + u(t, h(t)) + a(t, h(t)) \odot z(t, h_t), \ t \geq t_0,
\end{equation}
and $h_{t_0} = \phi \in C$ for some initial time $t_0 \in \mathbb{R}$, with the dependence on $x(t)$ viewed as dependence on $t$. 
Applying Theorem 3.7 to \citep{smith2011introduction}, we obtain the following result.
See App.~\ref{sect:appB} for a proof of this theorem.

\begin{theorem}[Existence and uniqueness of solution for continuous-time $\tau$-GRU] \label{thm_exist2main}
Let $t_0 \in \RR$ and $\phi \in C$ be given. There exists a unique solution $h(t) = h(t, \phi)$ of Eq. \eqref{eq_gendde}, defined on $[t_0 - \tau, t_0 + A]$ for any $A > 0$. In particular, the solution exists for all $t \geq t_0$, and
\begin{equation*}
    \| h_t(\phi) - h_t(\psi) \| \leq \| \phi - \psi \| e^{K(t-t_0)},
\end{equation*}
for all $t \geq t_0$, where $K = 1 + \|W_1\| +  \|W_2\| + \|W_4\|/4$.
\end{theorem}

Theorem \ref{thm_exist2main} guarantees that the continuous-time $\tau$-GRU, as a functional differential equation, has a well-defined unique solution that does not blow up in finite time. 

\subsection{The Delay Mechanism in \texorpdfstring{$\tau$}{tau}-GRUs Can Help Improve Long-Term Dependencies}

RNNs suffer from the problem of vanishing and exploding gradients, which leads to the problem of long-term dependencies.
Although the gating mechanisms could mitigate the problem to some extent, the delays introduced in $\tau$-GRUs can further 
help reduce the sensitivity to long-term dependencies.

To understand the reason for this, we consider how gradients are computed using the backpropagation through time (BPTT) algorithm~\citep{pascanu2013difficulty}. 
BPTT involves the two stages of unfolding the network in time and backpropagating the training error through the unfolded network. When $\tau$-GRUs are unfolded in time, the delays in the hidden state will appear as jump-ahead connections (buffers) in the unfolded network. These buffers provide a shorter path for propagating gradient information, and therefore reducing the sensitivity of the network to long-term dependencies. This intuition is also used to explain the behavior of the NARX RNNs in \citep{lin1996learning}. 

We use a simplified setting to make this intuition precise. 
See App.~\ref{sect:appC} for a proof of this proposition.
We also provide results (bounds for the gradient norm) and discussions for $\tau$-GRU (in App.~\ref{app:gradbound}, see Proposition \ref{app_prop}).
 
\begin{proposition} \label{prop_delaymain}
Consider the linear time-delayed RNN, with hidden states described by the update equation:
\begin{equation}
    h_{n+1} = Ah_n + Bh_{n-m} + Cu_n, \ n=0,1,\dots,
\end{equation}
and $h_n = 0$ for $n=-m, -m+1, \dots, 0$ with $m > 0$. 
Then, assuming that $A$ and $B$ commute, we have:
\begin{equation}
    \frac{\partial h_{n+1}}{\partial u_i} = A^{n-i} C, 
\end{equation}
for $n=0,1, \dots, m$, $i=0,\dots, n$, and 
\begin{align*}
    \frac{\partial h_{m+1+j}}{\partial u_i} &= A^{m+j-i} C + \delta_{i,j-1} BC + 2 \delta_{i,j-2} ABC \nonumber \\
    &\ \ \ \ + 3 \delta_{i,j-3} A^2 B C + \cdots + j \delta_{i,0} A^{j-1} B C, 
\end{align*}
for $j = 1,2,\dots, m+1$, $i=0,1,\dots, m+j$, where $\delta_{i,j}$ denotes the Kronecker delta.
\end{proposition}

We remark that the commutativity assumption is not necessary. It is used here only to simplify the expression for the gradients. An analogous formula for the gradients can be derived without such an assumption, at the cost of displaying more complicated formulae. 

From Proposition \ref{prop_delaymain}, we see that the presence of the delay allows the model to place more emphasis on the gradients due to input information in the past (as can also be seen in Eq. \eqref{eq_linear} in the proof of the proposition, where additional terms dependent on $B$ appear in the coefficients in front of the previous inputs). In particular, if $\|A\| < 1$ and $B=0$, then the gradients decay exponentially as $i$ becomes large. Introducing the delay term ($B\neq 0$) dampens the exponential decay by perturbing the gradients of hidden states (dependent on the delay parameter $m$)  with respect to past inputs with non-zero values, thus lessening the issue of vanishing gradients.

Similar qualitative conclusions can also be drawn for $\tau$-GRU (see Proposition \ref{app_prop} and the discussions in App. \ref{app:gradbound}). Therefore, we expect that these networks would be able to deal with long-term dependencies more effectively than the counterpart models without delays. 

\section{EXPERIMENTAL RESULTS}
\label{sect:exp}

In this section, we consider several tasks to demonstrate the performance of $\tau$-GRU compared to existing RNN models (see additional experiments in App.~\ref{sect:appD}).
We use standard protocols for training and validation sets for parameter tuning (see App.~\ref{app_tuningparam} for details and a study of sensitivity to random initialization).

\begin{figure*}[!t]
	\centering
	\begin{subfigure}[b]{0.49\textwidth}
		\centering
		\includegraphics[width=0.99\textwidth]{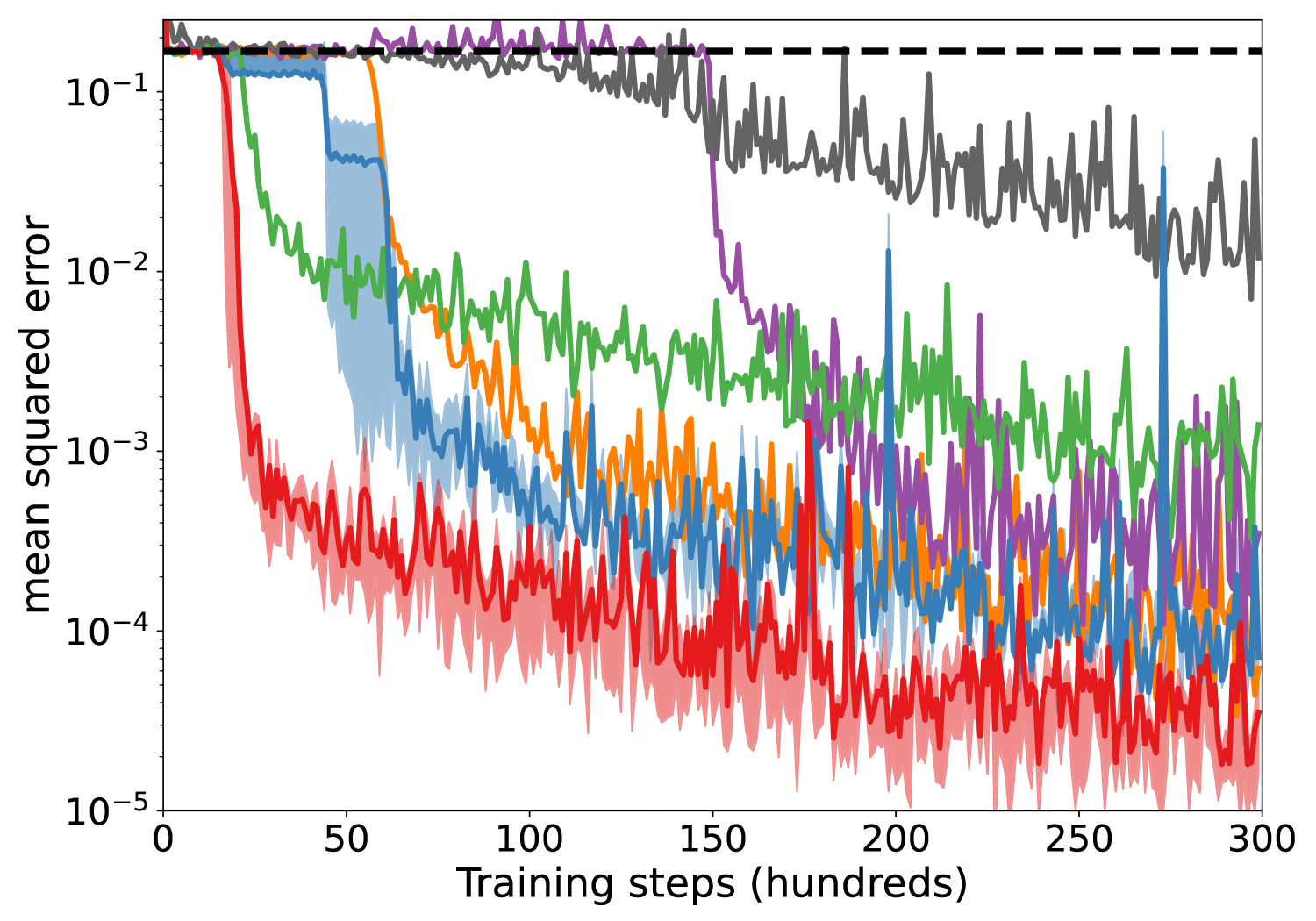}
		\caption{Sequence length $N=2000$.}
		\label{fig:adding_task_2000}
	\end{subfigure}
 ~
	\begin{subfigure}[b]{0.49\textwidth}
		\centering
		\includegraphics[width=0.99\textwidth]{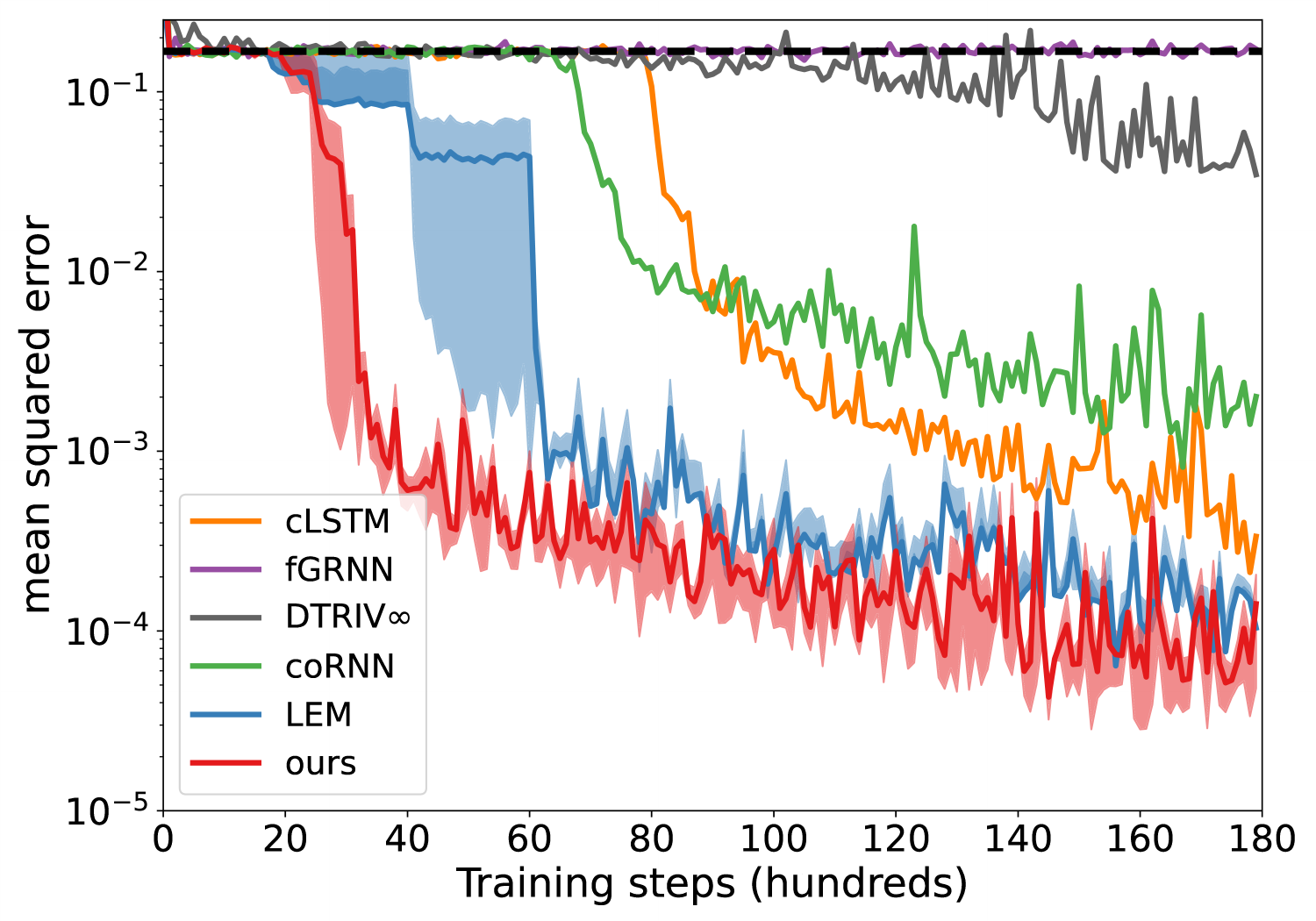}
		\caption{Sequence length $N=5000$.}
		\label{fig:adding_task_5000}
	\end{subfigure}	

	\caption{Results for the adding task. We show the one standard deviation bands for LEM and our $\tau$-GRU. On average, $\tau$-GRU converges faster, and obtains a lower MSE on the adding task. }
	\label{fig:adding_task}
\end{figure*}

\paragraph{The Adding Task.}
The adding task, proposed by \citet{hochreiter1997long}, tests a model's ability to learn long-term dependencies. The inputs are two stacked random vectors $u$ and $v$ of length $N$. Elements of $u$ are drawn from $\mathcal{U}(0,1)$, while $v$ has two non-zero elements (both set to 1) at random locations $i \in {1, \dots, \lfloor \frac{N}{2} \rfloor}$ and $j \in {\lceil \frac{N}{2} \rceil, \dots, N}$. The target value is  $\sum(u \odot v)$, i.e., the sum of the two elements in $u$ corresponding to the non-zero entries in $v$.

Following~\citet{rusch2022long}, we consider two challenging settings with very long input sequences $N={2000,5000}$. Figure~\ref{fig:adding_task} shows results for our $\tau$-GRU compared to several state-of-the-art RNN models designed for long-term dependency tasks, such as LEM~\citep{rusch2022long}, coRNN~\citep{rusch2021coupled}, DTRIV$\infty$\citep{lezcano2019trivializations}, fastGRNN\citep{kusupati2018fastgrnn}, and LSTM with chrono initialization~\citep{tallec2018can}. DTRIV$\infty$ and fastGRNN perform poorly in both cases, while our $\tau$-GRU converges faster and performs better.

\paragraph{Human Activity Recognition: HAR-2.}
Next, we evaluate our model's performance on human activity recognition using the HAR dataset~\citep{anguita2012human}, which includes accelerometer and gyroscope measurements from a Samsung Galaxy S3 smartphone tracking six activities performed by 30 volunteers aged 19-48. The sequences are divided into shorter segments of length $N=128$, with the raw measurements summarized by 9 features per time step. The HAR-2 dataset~\citep{kusupati2018fastgrnn} groups the activities into two categories. We use $7,352$ sequences for training, $900$ for validation, and $2,947$ for testing.

Our $\tau$-GRU outperforms traditional gated architectures on this task, as shown in Table~\ref{tab:results_har2}. The most competitive model is coRNN~\citep{rusch2021coupled}, achieving $97.2\%$ test accuracy with just $9$k parameters. LEM~\citep{rusch2022long} achieves $97.1\%$ with the same number of parameters as our $\tau$-GRU.

\begin{table}[!hb]
     \caption{Results for HAR2 task.}
	\label{tab:results_har2}
	\scalebox{0.69}{
		\begin{tabular}{l c  c ccccccc}
			\toprule
			Model & Test Acc. ($\%$) &  \# units & \# param \\			
			\midrule 
        GRU~\citep{kusupati2018fastgrnn} & 93.6 &  75 &   {19k}\\
        LSTM~\citep{Kag2020RNNs} & 93.7 &  64 &   {19k}\\
        FastRNN~\citep{kusupati2018fastgrnn} &  94.5 &  80 &  7k\\
        FastGRNN~\citep{kusupati2018fastgrnn} &  95.6 &  80 &  7k\\
        AsymRNN~\citep{Kag2020RNNs} &  93.2 &  120 & 8k\\
        iRNN~\citep{Kag2020RNNs} &  96.4 &  64 &  4k\\
        DIRNN~\citep{zhang2021deep} & 96.5 & 64 & -\\
        coRNN~\citep{rusch2021coupled} &  97.2 &  64 &  9k\\
        LipschitzRNN & 95.4 & 64 & 9k\\
        LEM & 97.1  &  64 &   {19k}\\        
		\midrule
		\textbf{$\tau$-GRU (ours)} & \textbf{97.4} & 64 & {19k} \\
			\bottomrule
	\end{tabular}}
\end{table}

\paragraph{Sentiment Analysis: IMDB.}
Here, we test the expressiveness of our proposed model on the sentiment analysis task using the IMDB dataset~\citep{maas2011learning}. This dataset consists of $50$k movie reviews, each labeled with a positive or negative sentiment, with an average length of 240 words. The dataset is evenly split into a training set and a test set, with 15\% of the training data used for validation. After standard preprocessing, we embed the data using a pretrained GloVe model~\citep{pennington2014glove}, restricting the dictionary to $25$k words. Our $\tau$-GRU achieves higher test accuracy than LSTM, GRU, continuous-time coRNN~\citep{rusch2021coupled}, and LEM~\citep{rusch2022long}, as shown in Table~\ref{tab:results_imbd}.

\begin{table}[!ht]
	\caption{Results for the IMDB  task.}
	\label{tab:results_imbd}
	\centering
	\scalebox{0.69}{
		\begin{tabular}{l c  c ccccccc}
			\toprule
			Model & Test Acc. ($\%$) &  \# units & \# param \\			
			\midrule 
            LSTM~\citep{campos2018skip} & 86.8 & 128 & {220k} \\
            Skip LSTM~\citep{campos2018skip} & 86.6 & 128 & {220k} \\
            GRU~\citep{campos2018skip} & 86.2 & 128 & 164k \\
            Skip GRU~\citep{campos2018skip} & 86.6 & 128 & 164k \\
            ReLU GRU~\citep{dey2017gate} & 84.8 & 128 & 99k \\
            coRNN~\citep{rusch2021coupled} & 87.4 & 128 & 46k \\
            LEM & 88.1 & 128 & {220k} \\
			\midrule
			\textbf{$\tau$-GRU (ours)} & \textbf{88.7} & 128 & {220k} \\
			\bottomrule
	\end{tabular}}
\end{table}

\begin{table*}[!ht]
	\caption{Test accuracies on sMNIST, psMNIST, sCIFAR, and nCIFAR.}
	\label{tab:image_mnist}
	\centering
		\scalebox{0.8}{
	\begin{tabular}{lcccc|cccc}
		\toprule
		Model& sMNIST & psMNIST  &  \# units & \# parms & sCIFAR & nCIFAR  &  \# units & \# parms \\
		\midrule
		LSTM~\citep{kag2021time} &  97.8 & 92.6 & 128 & {68k} & 59.7 & 11.6 & 128 & 69k / 117k \\
		r-LSTM~\citep{trinh2018learning} & 98.4  & 95.2 & - & 100K & 72.2 & - & - &  101k / -\\
		chrono-LSTM~\citep{rusch2022long} & 98.9 & 94.6  & 128 & {68k} & - & 55.9 & 128 & - / 116k\\
		Antisym. RNN~\citep{chang2018antisymmetricrnn} &  98.0 & 95.8  & 128 & 10k & 62.2 & 54.7 & 256 & 37k / 37k\\
		Lipschitz RNN~\citep{erichson2020lipschitz} & 99.4 & 96.3 & 128 & 34k & 64.2 & 59.0 & 256 &  134k / 158k\\
		expRNN~\citep{lezcano2019cheap} &  98.4 &  96.2&  360 & {68k} & - & 49.0  & 128 & - / 47k\\
		iRNN~\citep{Kag2020RNNs} & 98.1 & 95.6 & 128 & 8k & - & 54.5 & 128 & - / 12k \\
		TARNN~\citep{kag2021time} & 98.9  & 97.1 & 128 & {68k} & - & 59.1 & 128 & - /  100K \\
		
		Dilated GRU~\citep{chang2017dilated} & 99.2  & 94.6 & - & {130k} & - & - & - & - / - \\
		
		coRNN~\citep{rusch2021coupled} & 99.3 & {96.6}  & 128  & 34k & - & 59.0 & 128 & - / 46k\\
		{LEM}~\citep{rusch2022long} & \textbf{99.5} & {96.6}  & 128 & {68k}& - & 60.5 & 128 & - / 117k\\
  
		\midrule
		Delay GRU (Eq.~\eqref{eq_simpleDRNN}) & 98.7  &  94.1 & 128 & 51k & 56.1 & 53.7 & 128  & 52k / 75k  \\
		
		\textbf{$\tau$-GRU (ours)} & {99.4} & 97.3  & 128 & {68k} & \textbf{74.9} & \textbf{62.7} & 128 & 69k / 117k \\
		\bottomrule
	\end{tabular}}
\end{table*}

\paragraph{Sequential Image Classification.}
We evaluate the long-term dependency learning capabilities of RNNs on four image classification datasets: sequential MNIST (sMNIST), permuted sMNIST (psMNIST), sequential CIFAR-10 (sCIFAR), and noise-padded CIFAR-10 (nCIFAR).
For the sMNIST and psMNIST tasks, $N=784$ pixels of each image are sequentially presented to the RNN, with psMNIST using a fixed random permutation. The sCIFAR task has a sequence length of $N=1024$, with each element being a 3D vector representing the pixels for each color channel. The nCIFAR task uses a sequence of $N=1000$ with 968 noisy elements of dimension $96$. 

\begin{figure}[!b] 
\vspace{-0.1cm}
        \centering
		\includegraphics[width=0.49\textwidth]{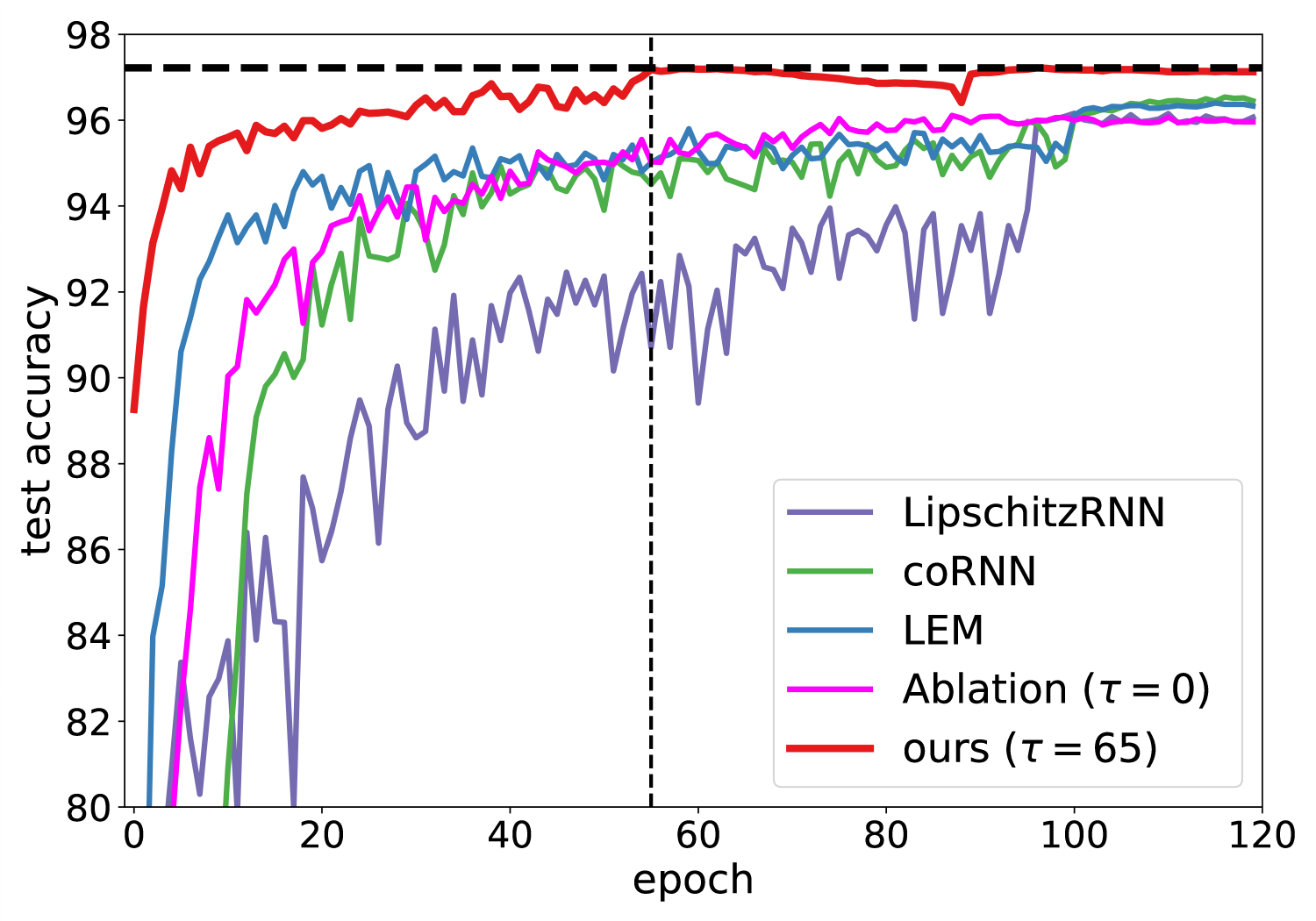}
        \vspace{-0.8cm}
        \caption{Test accuracy for psMNIST. }
		\label{fig:psmnist_acc}
\end{figure}

$\tau$-GRU outperforms other RNNs on the psMNIST, sCIFAR, and nCIFAR tasks, demonstrating a notable advantage in the CIFAR tasks through the proposed weighted time-delay feedback mechanism. As shown in Figure~\ref{fig:psmnist_acc}, our model converges faster than other continuous-time models, such as LEM, coRNN, and LipschitzRNN, requiring significantly fewer epochs to reach peak performance.

\begin{table}[!h]
	\caption{Results for the ENSO task.}
	\label{tab:results_ENSO}
	\centering
	\scalebox{0.69}{
		\begin{tabular}{l c  c  ccccccc}
			\toprule
			Model & MSE ($\times 10^{-2}$) &  \# units & \# parameters \\			
			\midrule 
			Vanilla RNN & 0.45 & 16 & 0.3k \\ 
			LSTM & 0.92 & 16 & 1.2k \\
			GRU & 0.53 & 16 & 0.9k \\
			Lipschitz RNN & 10.6 & 16 & 0.6k \\
			coRNN & 4.00 & 16 & 0.6k \\
			LEM & 0.31 & 16 & 1.2k \\
			\midrule
		  ablation ($\alpha = 0$) & 0.31 & 16 & 0.6k \\
		    ablation ($\beta = 0$) & 0.38 & 16 & 0.9k \\
			\midrule
			\textbf{$\tau$-GRU (ours)} & \textbf{0.17} & 16 & 1.2k \\
    	\bottomrule
	\end{tabular}}
\end{table}        

\paragraph{Learning Climate Dynamics.}

We consider the task of learning the dynamics of the DDE model for El Ni\~{n}o Southern Oscillation (ENSO)  of \citep{falkena2019derivation} (see Eq. (46) there). It models the sea surface temperature $T$ in the eastern Pacific Ocean as:
\begin{equation}
    \dot{T} = T-T^3 - c T(t-\delta) (1-\gamma T^2(t-\delta)), \ t > \delta,
\end{equation}
where $\gamma < 1$, $c > 0$, with $T(t)$ satisfying $\dot{T} = T-T^3 - c T(0) (1-\gamma T(0)^2)$ with $T(0) \sim \rm{Unif}(0,1)$ for $t \in [0, \delta]$.   For data generation, we follow \cite{falkena2019derivation}, and choose $c = 0.93$, $\gamma = 0.49$ and $\delta = 4.8$. We use the classical Runge-Kutta  method (RK4) to numerically integrate the system from $t=0$ to $t=400$ using a step-size of 0.1. The training and testing samples are the time series (of length 2000) generated by the RK4 scheme on the interval $[200,400]$ for different realizations of $T(0)$.  

Table \ref{tab:results_ENSO} shows that our model ($\alpha=\beta=1$, $\tau = 20$) is more effective in learning the ENSO dynamics compared to other models. We also see that the predictive performance deteriorates without using the appropriate combination of standard and delay recurrent units (setting either $\alpha$ or $\beta$ to zero).

\paragraph{Frequency Classification.}

Next, we consider the frequency classification task introduced by \cite{moreno2024rough}. 
The dataset consists of $N$ time series samples in 100 distinct frequency classes, one for each frequency $f_j$, linearly spaced between $1$ and $2^{12}$:
\begin{equation}
f_j = 1 + (j-1) \frac{2^{12} - 1}{99}, \ \ j=1,\dots, 100,
\end{equation}
which are used to generate the length-$L$  cosine signals $X_j(t_n) = \cos(2 \pi f_j t_n) + \sigma \epsilon_j(t_n)$, where $t_n = n \Delta t$, $n=0,1,\dots, L-1$, $\sigma_j \geq 0$,  and the $\epsilon_j(t_n)$ are i.i.d. standard Gaussians. 
We generate 1000 samples (that is, $N=10$) across 100 distinct classes, with each of them sampled uniformly at $1000$ time steps in the interval $[0,1]$. We consider both the noise-free $(\sigma = 0$) and the more challenging noisy case $(\sigma = 0.1)$.

Table~\ref{tab:frequency_results} summarizes the classification accuracy obtained for different models. RNNs show strong performance on the noise-free data set, but they significantly degrade on the noisy data set. The SSM S4 model~\citep{gu2021efficiently} struggles to clearly discriminate frequency classes, showing a limited ability to capture nonlinear frequency dependencies within linear dynamics. This limitation arises since linear time-invariant systems in SSMs function effectively as linear filters, and thus primarily scale Fourier modes rather than distinguishing between them through nonlinear recurrence. In contrast, our $\tau$-GRU  outperforms all other methods, achieving perfect accuracy (100\%) in the noise-free scenario and near-perfect (99.1\%) accuracy in the noisy scenario. Furthermore, $\tau$-GRU converges quickly, attaining 100\% test accuracy within only 3 epochs in the noise-free scenario, while baseline methods typically require between 11 and 45 epochs to converge. Similarly, in the noisy setting, our model reaches 99\% accuracy after merely 15 epochs, substantially faster than other methods.

\begin{table}[!ht]
    \centering
    \caption{Accuracy on the frequency classification task.}
    \label{tab:frequency_results}
    \scalebox{0.69}{
    \begin{tabular}{lcc}
        \toprule
        \textbf{Model} & \textbf{No noise} & \textbf{With noise} \\
        \midrule
        Tanh-RNN & 97.1\% & 35.6\% \\
        LSTM & 100.0\% & 39.4\% \\
        LSTM (w/o forget gate) & 99.0\% & 19.4\% \\
        LEM & 96.0\% & 54.1\% \\
        SSM-S4D (1 layer) & 67.5\% & 66.4\% \\
        SSM-S4D (4 layers) & 68.9\% & 67.2\% \\
        GRU (no delay, ablation) & 95.0\% & 57.7\% \\
        \midrule
        \textbf{GRU (with delay, ours)} & \textbf{100.0\%} & \textbf{99.1\%} \\
        \bottomrule
    \end{tabular}}
\end{table}

\paragraph{Ablation Study using psMNIST.}

We used the psMNIST task to perform an ablation study. To do so, we consider the following model
$$\bh_{t+1} = (1-\bg_t) \odot \bh_t + \bg_t \odot (\beta \cdot \bu_t + \alpha \cdot \ba_t \odot \bz_t),$$
where $\alpha\in [0,1]$ and $\beta\in [0,1]$ are constants that can be used to control the effect of different components. We are interested in the cases where $\alpha$ and $\beta$ are either 0 or 1, i.e., a component is switched off or on. 
%
Table~\ref{table:ablation} shows the results for different ablation configurations. By setting $\alpha=0$ we yield a simple gated RNN.
Second, for $\beta=0$, we yield a $\tau$-GRU without instantaneous dynamics. Third, we show how different values of $\tau$ affect the performance. Setting $\tau=0$ leads to a $\tau$-GRU without time-delay feedback.
We also show that a model without the weighting function $\ba_t$ is not able to achieve peak performance.

\begin{table}[!t]
    \centering
 	\caption{Ablation study on psMNIST.}
	\label{table:ablation}
	\scalebox{0.69}{
		\begin{tabular}{l c  c  ccccccc}
			\toprule
			Model & $\alpha$ &  $\beta$ & $\tau$ & $\ba_t$ & Accuracy (\%) \\			
			\midrule 
			ablation & 0 & 1 & - & yes & 94.6 \\
			ablation & 1 & 0 & 65 & yes &94.9 \\
			\midrule
			ablation & 1 & 1 & 0 & yes & 95.1 \\
			ablation & 1 & 1 & 20 & yes & 96.4 \\
			ablation & 1 & 1 & 65 & no & 96.8 \\
			\midrule
			\textbf{$\tau$-GRU (ours)} & 1 & 1 & 65 & yes &\textbf{97.3}\\
			\bottomrule 
	\end{tabular}}        
\end{table}

\begin{figure}[!t]
    \centering
	\includegraphics[width=0.49\textwidth]{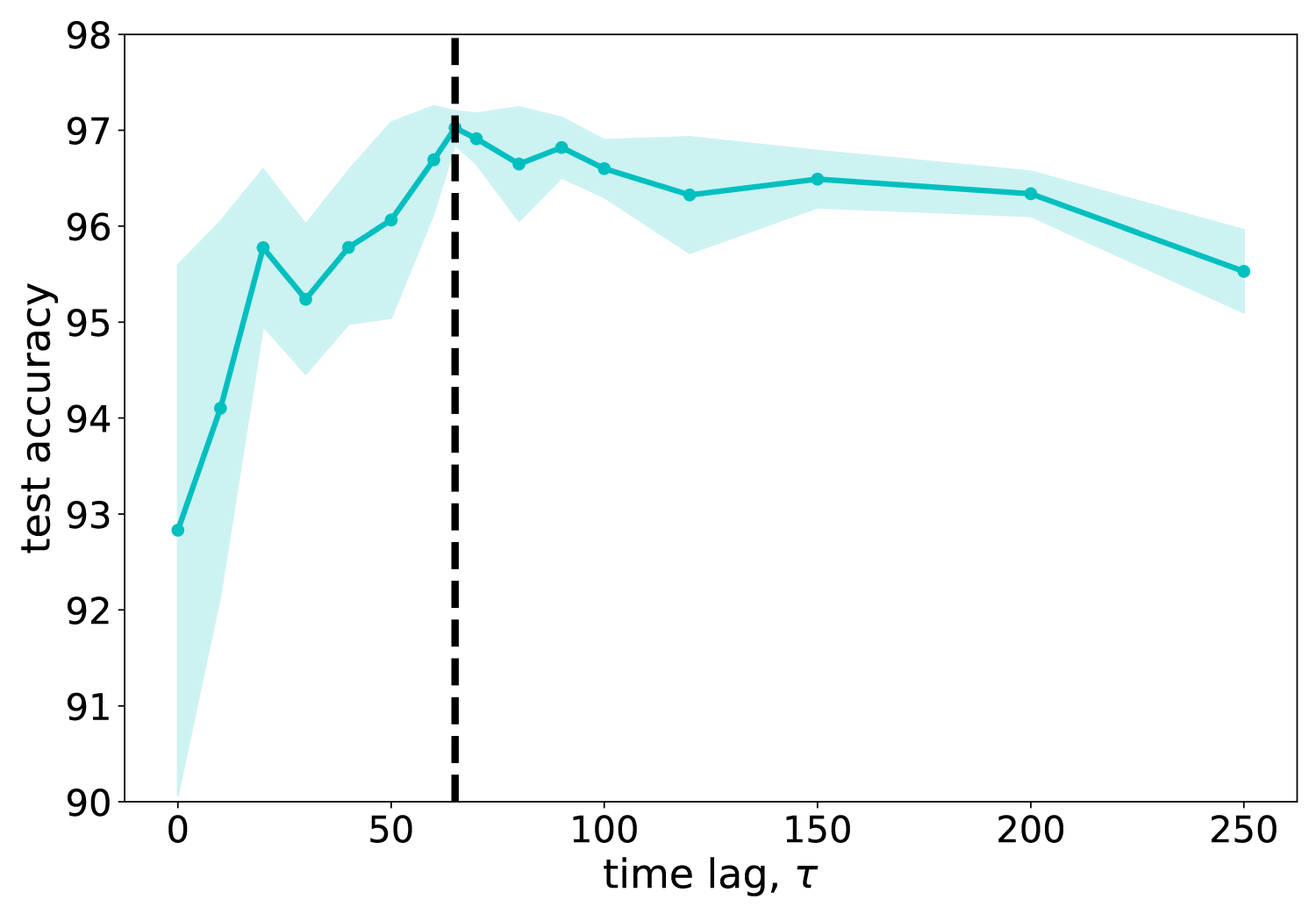}
     \vspace{-0.8cm}
	\caption{Sensitivity analysis of $\tau$-GRU on psMNIST.  The green envelope represent $\pm 1$ s.d. around the mean.}
	\label{fig:tau_ablation_main}
\end{figure}

Figure~\ref{fig:tau_ablation_main} demonstrates the effect of $\tau$. The performance of $\tau$-GRU is increasing as a function of $\tau$ and peaks around $\tau=65$.  
It can be seen that the performance in the range $50$-$150$ is relatively constant for this task. Thus, the model is relatively insensitive as long as $\tau$ is sufficiently large, but not too large. Performance is starting to decrease for $\tau>150$.  Since $\tau$ takes discrete values, tuning is easier compared to continuous tuning parameters, for example, parameters used by LEM~\citep{rusch2022long}, coRNN~\citep{rusch2021coupled}, or LipschitzRNN~\citep{erichson2020lipschitz}.

\section{CONCLUSION}

Starting from a continuous-time formulation, we derive a discrete-time gated recurrent unit with  delay, $\tau$-GRU. We also provide intuition and theoretical analysis to understand how the proposed delay term can improve the modeling of long-term dependencies. 
Importantly, we demonstrate the superior performance of $\tau$-GRU in improving the long-range modeling capability of existing RNN models in a wide range of challenging tasks, from sequential image classification to learning nonlinear dynamics, given a comparable number of trainable parameters. Although there exist several other more sophisticated models, such as state-space models (SSMs) \citep{gu2021combining, gu2021efficiently}, these models may not be optimal for tasks such as learning nonlinear dynamics in the {\it small data regime}, which is quite common in many scientific applications (see App. \ref{sect_AppH}).

One limitation of $\tau$-GRU is that we consider only recurrent units with a single delay instead of multiple delays, which limits the potential of our architecture. We restrict ourselves to the single delay case to enable tractable analysis and experimentation.
We shall leave the extension to include distributed delay mechanisms for future work. It would also be interesting to study noise-injected versions of $\tau$-GRU to improve the trade-offs between accuracy and robustness to data perturbations \cite{lim2021noisy, lim2021nfm, erichson2022noisymix}.

\subsubsection*{Acknowledgments} 
NBE would like to acknowledge NSF, under Grant No. 2319621, and the U.S.
Department of Energy, under Contract Number DE-AC02-05CH11231 and DE-AC02-05CH11231,
for providing partial support of this work. 
SHL acknowledges support from the Wallenberg Initiative on Networks and Quantum
Information (WINQ), the Swedish Research Council (VR/2021-03648), and the resources provided by the National Academic Infrastructure
for Supercomputing in Sweden (NAISS), partially funded by the Swedish Research Council
through grant agreement no. 2022-06725 (NAISS 2024/5-269).

\bibliography{references} 

\begin{thebibliography}{}

\bibitem[Anguita et~al., 2012]{anguita2012human}
Anguita, D., Ghio, A., Oneto, L., Parra, X., and Reyes-Ortiz, J.~L. (2012).
\newblock Human activity recognition on smartphones using a multiclass
  hardware-friendly support vector machine.
\newblock In {\em International Workshop on Ambient Assisted Living}, pages
  216--223. Springer.

\bibitem[Arjovsky et~al., 2016]{arjovsky2016unitary}
Arjovsky, M., Shah, A., and Bengio, Y. (2016).
\newblock Unitary evolution recurrent neural networks.
\newblock In {\em International Conference on Machine Learning}, pages
  1120--1128. PMLR.

\bibitem[Asl and Ulsoy, 2003]{asl2003analysis}
Asl, F.~M. and Ulsoy, A.~G. (2003).
\newblock Analysis of a system of linear delay differential equations.
\newblock {\em J. Dyn. Sys., Meas., Control}, 125(2):215--223.

\bibitem[Azencot et~al., 2021]{azencot2021differential}
Azencot, O., Erichson, N.~B., Ben-Chen, M., and Mahoney, M.~W. (2021).
\newblock A differential geometry perspective on orthogonal recurrent models.
\newblock {\em arXiv preprint arXiv:2102.09589}.

\bibitem[Baldi and Atiya, 1994]{baldi1994delays}
Baldi, P. and Atiya, A.~F. (1994).
\newblock How delays affect neural dynamics and learning.
\newblock {\em IEEE Transactions on Neural Networks}, 5(4):612--621.

\bibitem[Barron, 1993]{barron1993universal}
Barron, A.~R. (1993).
\newblock Universal approximation bounds for superpositions of a sigmoidal
  function.
\newblock {\em IEEE Transactions on Information theory}, 39(3):930--945.

\bibitem[B{\'e}lair, 1993]{belair1993stability}
B{\'e}lair, J. (1993).
\newblock Stability in a model of a delayed neural network.
\newblock {\em Journal of Dynamics and Differential Equations}, 5(4):607--623.

\bibitem[Campbell, 2007]{campbell2007time}
Campbell, S.~A. (2007).
\newblock Time delays in neural systems.
\newblock In {\em Handbook of Brain Connectivity}, pages 65--90. Springer.

\bibitem[Campos et~al., 2018]{campos2018skip}
Campos, V., Jou, B., Gir{\'o}-i Nieto, X., Torres, J., and Chang, S.-F. (2018).
\newblock Skip {R}{N}{N}: Learning to skip state updates in recurrent neural
  networks.
\newblock In {\em International Conference on Learning Representations}.

\bibitem[Chang et~al., 2018]{chang2018antisymmetricrnn}
Chang, B., Chen, M., Haber, E., and Chi, E.~H. (2018).
\newblock Antisymmetric {R}{N}{N}: A dynamical system view on recurrent neural
  networks.
\newblock In {\em International Conference on Learning Representations}.

\bibitem[Chang et~al., 2017]{chang2017dilated}
Chang, S., Zhang, Y., Han, W., Yu, M., Guo, X., Tan, W., Cui, X., Witbrock, M.,
  Hasegawa-Johnson, M.~A., and Huang, T.~S. (2017).
\newblock Dilated recurrent neural networks.
\newblock {\em Advances in neural information processing systems}, 30.

\bibitem[Chen et~al., 2018]{chen2018neural}
Chen, R.~T., Rubanova, Y., Bettencourt, J., and Duvenaud, D.~K. (2018).
\newblock Neural ordinary differential equations.
\newblock {\em Advances in Neural Information Processing Systems}, 31.

\bibitem[Cho et~al., 2014]{cho2014properties}
Cho, K., Van~Merri{\"e}nboer, B., Bahdanau, D., and Bengio, Y. (2014).
\newblock On the properties of neural machine translation: Encoder-decoder
  approaches.
\newblock {\em arXiv preprint arXiv:1409.1259}.

\bibitem[Dey and Salem, 2017]{dey2017gate}
Dey, R. and Salem, F.~M. (2017).
\newblock Gate-variants of gated recurrent unit ({G}{R}{U}) neural networks.
\newblock In {\em 2017 IEEE 60th International Midwest Symposium on Circuits
  and Systems (MWSCAS)}, pages 1597--1600. IEEE.

\bibitem[Erichson et~al., 2020]{erichson2020lipschitz}
Erichson, N.~B., Azencot, O., Queiruga, A., Hodgkinson, L., and Mahoney, M.~W.
  (2020).
\newblock Lipschitz recurrent neural networks.
\newblock In {\em International Conference on Learning Representations}.

\bibitem[Erichson et~al., 2022]{erichson2022noisymix}
Erichson, N.~B., Lim, S.~H., Utrera, F., Xu, W., Cao, Z., and Mahoney, M.~W.
  (2022).
\newblock Noisy{M}ix: Boosting robustness by combining data augmentations,
  stability training, and noise injections.
\newblock {\em arXiv preprint arXiv:2202.01263}.

\bibitem[Erneux, 2009]{erneux2009applied}
Erneux, T. (2009).
\newblock {\em Applied Delay Differential Equations}, volume~3.
\newblock Springer.

\bibitem[Falkena et~al., 2019]{falkena2019derivation}
Falkena, S.~K., Quinn, C., Sieber, J., Frank, J., and Dijkstra, H.~A. (2019).
\newblock Derivation of delay equation climate models using the
  {M}ori-{Z}wanzig formalism.
\newblock {\em Proceedings of the Royal Society A}, 475(2227):20190075.

\bibitem[Gu and Dao, 2023]{gu2023mamba}
Gu, A. and Dao, T. (2023).
\newblock Mamba: Linear-time sequence modeling with selective state spaces.
\newblock {\em arXiv preprint arXiv:2312.00752}.

\bibitem[Gu et~al., 2020]{gu2020hippo}
Gu, A., Dao, T., Ermon, S., Rudra, A., and R{\'e}, C. (2020).
\newblock Hippo: Recurrent memory with optimal polynomial projections.
\newblock {\em Advances in Neural Information Processing Systems},
  33:1474--1487.

\bibitem[Gu et~al., 2021a]{gu2021efficiently}
Gu, A., Goel, K., and R{\'e}, C. (2021a).
\newblock Efficiently modeling long sequences with structured state spaces.
\newblock {\em arXiv preprint arXiv:2111.00396}.

\bibitem[Gu et~al., 2021b]{gu2021combining}
Gu, A., Johnson, I., Goel, K., Saab, K., Dao, T., Rudra, A., and R{\'e}, C.
  (2021b).
\newblock Combining recurrent, convolutional, and continuous-time models with
  linear state space layers.
\newblock {\em Advances in Neural Information Processing Systems}, 34:572--585.

\bibitem[Hale and Lunel, 2013]{hale2013introduction}
Hale, J.~K. and Lunel, S. M.~V. (2013).
\newblock {\em Introduction to Functional Differential Equations}, volume~99.
\newblock Springer Science \& Business Media.

\bibitem[Hasani et~al., 2022a]{hasani2022closed}
Hasani, R., Lechner, M., Amini, A., Liebenwein, L., Ray, A., Tschaikowski, M.,
  Teschl, G., and Rus, D. (2022a).
\newblock Closed-form continuous-time neural networks.
\newblock {\em Nature Machine Intelligence}, pages 1--12.

\bibitem[Hasani et~al., 2022b]{hasani2022liquid}
Hasani, R., Lechner, M., Wang, T.-H., Chahine, M., Amini, A., and Rus, D.
  (2022b).
\newblock Liquid structural state-space models.
\newblock {\em arXiv preprint arXiv:2209.12951}.

\bibitem[Hochreiter and Schmidhuber, 1997]{hochreiter1997long}
Hochreiter, S. and Schmidhuber, J. (1997).
\newblock Long short-term memory.
\newblock {\em Neural Computation}, 9(8):1735--1780.

\bibitem[Jaeger et~al., 2007]{jaeger2007optimization}
Jaeger, H., Luko{\v{s}}evi{\v{c}}ius, M., Popovici, D., and Siewert, U. (2007).
\newblock Optimization and applications of echo state networks with
  leaky-integrator neurons.
\newblock {\em Neural Networks}, 20(3):335--352.

\bibitem[Kag and Saligrama, 2021]{kag2021time}
Kag, A. and Saligrama, V. (2021).
\newblock Time adaptive recurrent neural network.
\newblock In {\em Proceedings of the IEEE/CVF Conference on Computer Vision and
  Pattern Recognition}, pages 15149--15158.

\bibitem[Kag et~al., 2020]{Kag2020RNNs}
Kag, A., Zhang, Z., and Saligrama, V. (2020).
\newblock {R}{N}{N}s incrementally evolving on an equilibrium manifold: A
  panacea for vanishing and exploding gradients?
\newblock In {\em International Conference on Learning Representations}.

\bibitem[Keane et~al., 2017]{keane2017climate}
Keane, A., Krauskopf, B., and Postlethwaite, C.~M. (2017).
\newblock Climate models with delay differential equations.
\newblock {\em Chaos: An Interdisciplinary Journal of Nonlinear Science},
  27(11):114309.

\bibitem[Kerg et~al., 2019]{kerg2019non}
Kerg, G., Goyette, K., Touzel, M.~P., Gidel, G., Vorontsov, E., Bengio, Y., and
  Lajoie, G. (2019).
\newblock Non-normal recurrent neural network ({nn{R}{N}{N}}): Learning long
  time dependencies while improving expressivity with transient dynamics.
\newblock In {\em Advances in Neural Information Processing Systems}, pages
  13591--13601.

\bibitem[Kidger et~al., 2020]{kidger2020neural}
Kidger, P., Morrill, J., Foster, J., and Lyons, T. (2020).
\newblock Neural controlled differential equations for irregular time series.
\newblock {\em Advances in Neural Information Processing Systems},
  33:6696--6707.

\bibitem[Kim, 1998]{kim1998time}
Kim, S.-S. (1998).
\newblock Time-delay recurrent neural network for temporal correlations and
  prediction.
\newblock {\em Neurocomputing}, 20(1-3):253--263.

\bibitem[Kusupati et~al., 2018]{kusupati2018fastgrnn}
Kusupati, A., Singh, M., Bhatia, K., Kumar, A., Jain, P., and Varma, M. (2018).
\newblock Fast{G}{R}{N}{N}: A fast, accurate, stable and tiny kilobyte sized
  gated recurrent neural network.
\newblock {\em Advances in Neural Information Processing Systems}, 31.

\bibitem[L-Casado and M-Rubio, 2019]{lezcano2019cheap}
L-Casado, M. and M-Rubio, D. (2019).
\newblock Cheap orthogonal constraints in neural networks: A simple
  parametrization of the orthogonal and unitary group.
\newblock {\em International Conference on Machine Learning}, pages 3794--3803.

\bibitem[Lang et~al., 1990]{lang1990time}
Lang, K.~J., Waibel, A.~H., and Hinton, G.~E. (1990).
\newblock A time-delay neural network architecture for isolated word
  recognition.
\newblock {\em Neural Networks}, 3(1):23--43.

\bibitem[Lepri et~al., 1994]{lepri1994high}
Lepri, S., Giacomelli, G., Politi, A., and Arecchi, F. (1994).
\newblock High-dimensional chaos in delayed dynamical systems.
\newblock {\em Physica D: Nonlinear Phenomena}, 70(3):235--249.

\bibitem[Lezcano~Casado, 2019]{lezcano2019trivializations}
Lezcano~Casado, M. (2019).
\newblock Trivializations for gradient-based optimization on manifolds.
\newblock {\em Advances in Neural Information Processing Systems}, 32.

\bibitem[Liao et~al., 2002]{liao2002delay}
Liao, X., Chen, G., and Sanchez, E.~N. (2002).
\newblock Delay-dependent exponential stability analysis of delayed neural
  networks: an {L}{M}{I} approach.
\newblock {\em Neural Networks}, 15(7):855--866.

\bibitem[Lim et~al., 2021a]{lim2021noisy}
Lim, S.~H., Erichson, N.~B., Hodgkinson, L., and Mahoney, M.~W. (2021a).
\newblock Noisy recurrent neural networks.
\newblock {\em Advances in Neural Information Processing Systems},
  34:5124--5137.

\bibitem[Lim et~al., 2021b]{lim2021nfm}
Lim, S.~H., Erichson, N.~B., Utrera, F., Xu, W., and Mahoney, M.~W. (2021b).
\newblock Noisy feature mixup.
\newblock {\em arXiv preprint arXiv:2110.02180}.

\bibitem[Lin et~al., 1996]{lin1996learning}
Lin, T., Horne, B.~G., Tino, P., and Giles, C.~L. (1996).
\newblock Learning long-term dependencies in {N}{A}{R}{X} recurrent neural
  networks.
\newblock {\em IEEE Transactions on Neural Networks}, 7(6):1329--1338.

\bibitem[Maas et~al., 2011]{maas2011learning}
Maas, A., Daly, R.~E., Pham, P.~T., Huang, D., Ng, A.~Y., and Potts, C. (2011).
\newblock Learning word vectors for sentiment analysis.
\newblock In {\em Proceedings of the 49th Annual Meeting of the Association for
  Computational Linguistics: Human Language Technologies}, pages 142--150.

\bibitem[Mackey and Glass, 1977]{mackey1977oscillation}
Mackey, M.~C. and Glass, L. (1977).
\newblock Oscillation and chaos in physiological control systems.
\newblock {\em Science}, 197(4300):287--289.

\bibitem[Marcus and Westervelt, 1989]{marcus1989stability}
Marcus, C. and Westervelt, R. (1989).
\newblock Stability of analog neural networks with delay.
\newblock {\em Physical Review A}, 39(1):347.

\bibitem[Moreno-Pino et~al., 2024]{moreno2024rough}
Moreno-Pino, F., Arroyo, {\'A}., Waldon, H., Dong, X., and Cartea, {\'A}.
  (2024).
\newblock Rough transformers: Lightweight continuous-time sequence modelling
  with path signatures.
\newblock {\em arXiv preprint arXiv:2405.20799}.

\bibitem[Orvieto et~al., 2023]{orvieto2023resurrecting}
Orvieto, A., Smith, S.~L., Gu, A., Fernando, A., Gulcehre, C., Pascanu, R., and
  De, S. (2023).
\newblock Resurrecting recurrent neural networks for long sequences.
\newblock {\em arXiv preprint arXiv:2303.06349}.

\bibitem[Park et~al., 2019]{park2019dynamic}
Park, J.~H., Lee, T.~H., Liu, Y., and Chen, J. (2019).
\newblock {\em Dynamic Systems with Time Delays: Stability and Control}.
\newblock Springer.

\bibitem[Pascanu et~al., 2013]{pascanu2013difficulty}
Pascanu, R., Mikolov, T., and Bengio, Y. (2013).
\newblock On the difficulty of training recurrent neural networks.
\newblock In {\em International Conference on Machine Learning}, pages
  1310--1318. PMLR.

\bibitem[Pennington et~al., 2014]{pennington2014glove}
Pennington, J., Socher, R., and Manning, C.~D. (2014).
\newblock Glove: Global vectors for word representation.
\newblock In {\em Proceedings of the 2014 conference on empirical methods in
  natural language processing (EMNLP)}, pages 1532--1543.

\bibitem[Pineda, 1988]{pineda1988dynamics}
Pineda, F.~J. (1988).
\newblock Dynamics and architecture for neural computation.
\newblock {\em Journal of Complexity}, 4(3):216--245.

\bibitem[Queiruga et~al., 2021]{queiruga2021stateful}
Queiruga, A., Erichson, N.~B., Hodgkinson, L., and Mahoney, M.~W. (2021).
\newblock Stateful ode-nets using basis function expansions.
\newblock {\em Advances in Neural Information Processing Systems},
  34:21770--21781.

\bibitem[Queiruga et~al., 2020]{queiruga2020continuous}
Queiruga, A.~F., Erichson, N.~B., Taylor, D., and Mahoney, M.~W. (2020).
\newblock Continuous-in-depth neural networks.
\newblock {\em arXiv preprint arXiv:2008.02389}.

\bibitem[Rusch and Mishra, 2021]{rusch2021coupled}
Rusch, T.~K. and Mishra, S. (2021).
\newblock Coupled oscillatory recurrent neural network (co{R}{N}{N}): An
  accurate and (gradient) stable architecture for learning long time
  dependencies.
\newblock In {\em International Conference on Learning Representations}.

\bibitem[Rusch et~al., 2022]{rusch2022long}
Rusch, T.~K., Mishra, S., Erichson, N.~B., and Mahoney, M.~W. (2022).
\newblock Long expressive memory for sequence modeling.
\newblock In {\em International Conference on Learning Representations}.

\bibitem[Sch{\"a}fer and Zimmermann, 2006]{schafer2006recurrent}
Sch{\"a}fer, A.~M. and Zimmermann, H.~G. (2006).
\newblock Recurrent neural networks are universal approximators.
\newblock In {\em International Conference on Artificial Neural Networks},
  pages 632--640.

\bibitem[Smith, 2011]{smith2011introduction}
Smith, H.~L. (2011).
\newblock {\em An Introduction to Delay Differential Equations with
  Applications to the Life Sciences}, volume~57.
\newblock Springer New York.

\bibitem[Smith et~al., 2022]{smith2022simplified}
Smith, J.~T., Warrington, A., and Linderman, S.~W. (2022).
\newblock Simplified state space layers for sequence modeling.
\newblock {\em arXiv preprint arXiv:2208.04933}.

\bibitem[Tallec and Ollivier, 2018]{tallec2018can}
Tallec, C. and Ollivier, Y. (2018).
\newblock Can recurrent neural networks warp time?
\newblock In {\em International Conference on Learning Representations}.

\bibitem[Tay et~al., 2022]{tay2022efficient}
Tay, Y., Dehghani, M., Bahri, D., and Metzler, D. (2022).
\newblock Efficient transformers: A survey.
\newblock {\em ACM Computing Surveys}, 55(6):1--28.

\bibitem[Trinh et~al., 2018]{trinh2018learning}
Trinh, T., Dai, A., Luong, T., and Le, Q. (2018).
\newblock Learning longer-term dependencies in {R}{N}{N}s with auxiliary
  losses.
\newblock In {\em International Conference on Machine Learning}, pages
  4965--4974. PMLR.

\bibitem[Vaswani et~al., 2017]{vaswani2017attention}
Vaswani, A., Shazeer, N., Parmar, N., Uszkoreit, J., Jones, L., Gomez, A.~N.,
  Kaiser, {\L}., and Polosukhin, I. (2017).
\newblock Attention is all you need.
\newblock {\em Advances in Neural Information Processing Systems}, 30.

\bibitem[Voelker et~al., 2019]{voelker2019legendre}
Voelker, A., Kaji{\'c}, I., and Eliasmith, C. (2019).
\newblock Legendre memory units: Continuous-time representation in recurrent
  neural networks.
\newblock {\em Advances in Neural Information Processing Systems}, 32.

\bibitem[Vorontsov et~al., 2017]{vorontsov2017orthogonality}
Vorontsov, E., Trabelsi, C., Kadoury, S., and Pal, C. (2017).
\newblock On orthogonality and learning recurrent networks with long term
  dependencies.
\newblock In {\em International Conference on Machine Learning}, pages
  3570--3578. PMLR.

\bibitem[Waibel et~al., 1989]{waibel1989phoneme}
Waibel, A., Hanazawa, T., Hinton, G., Shikano, K., and Lang, K.~J. (1989).
\newblock Phoneme recognition using time-delay neural networks.
\newblock {\em IEEE Transactions on Acoustics, Speech, and Signal Processing},
  37(3):328--339.

\bibitem[Wang and Xue, 2023]{wang2023state}
Wang, S. and Xue, B. (2023).
\newblock State-space models with layer-wise nonlinearity are universal
  approximators with exponential decaying memory.
\newblock {\em arXiv preprint arXiv:2309.13414}.

\bibitem[Warden, 2018]{warden2018speech}
Warden, P. (2018).
\newblock Speech commands: A dataset for limited-vocabulary speech recognition.
\newblock {\em arXiv preprint arXiv:1804.03209}.

\bibitem[Wisdom et~al., 2016]{wisdom2016full}
Wisdom, S., Powers, T., Hershey, J., Le~Roux, J., and Atlas, L. (2016).
\newblock Full-capacity unitary recurrent neural networks.
\newblock {\em Advances in Neural Information Processing Systems}, 29.

\bibitem[Xia et~al., 2021]{xia2021heavy}
Xia, H., Suliafu, V., Ji, H., Nguyen, T., Bertozzi, A., Osher, S., and Wang, B.
  (2021).
\newblock Heavy ball neural ordinary differential equations.
\newblock {\em Advances in Neural Information Processing Systems},
  34:18646--18659.

\bibitem[Yang et~al., 2014]{yang2014exponential}
Yang, Z., Zhou, W., and Huang, T. (2014).
\newblock Exponential input-to-state stability of recurrent neural networks
  with multiple time-varying delays.
\newblock {\em Cognitive Neurodynamics}, 8(1):47--54.

\bibitem[Yu et~al., 2024a]{yu2024tuning}
Yu, A., Lyu, D., Lim, S.~H., Mahoney, M.~W., and Erichson, N.~B. (2024a).
\newblock Tuning frequency bias of state space models.
\newblock {\em arXiv preprint arXiv:2410.02035}.

\bibitem[Yu et~al., 2024b]{yu2024hope}
Yu, A., Mahoney, M.~W., and Erichson, N.~B. (2024b).
\newblock Hope for a robust parameterization of long-memory state space models.
\newblock {\em arXiv preprint arXiv:2405.13975}.

\bibitem[Yu et~al., 2023]{yu2023robustifying}
Yu, A., Nigmetov, A., Morozov, D., Mahoney, M.~W., and Erichson, N.~B. (2023).
\newblock Robustifying state-space models for long sequences via approximate
  diagonalization.
\newblock {\em arXiv preprint arXiv:2310.01698}.

\bibitem[Zhang et~al., 2021]{zhang2021deep}
Zhang, Z., Wu, G., Li, Y., Yue, Y., and Zhou, X. (2021).
\newblock Deep incremental {R}{N}{N} for learning sequential data: A {L}yapunov
  stable dynamical system.
\newblock In {\em 2021 IEEE International Conference on Data Mining (ICDM)},
  pages 966--975. IEEE.

\bibitem[Zhu et~al., 2021]{zhu2021neural}
Zhu, Q., Guo, Y., and Lin, W. (2021).
\newblock Neural delay differential equations.
\newblock {\em arXiv preprint arXiv:2102.10801}.

\end{thebibliography}


\newpage

\onecolumn

\begin{center}
    {\Large \bf  Appendix}
\end{center}
\appendix

This Appendix is organized as follows. In Section \ref{sect:appA}, we provide illustrations to demonstrate the differences between the ODE and DDE dynamics driven by input in the scalar setting. In Section \ref{sect:appB}, we provide the proof of Theorem \ref{thm_exist2main}. In Section \ref{sect:appC}, we provide the proof of Proposition \ref{prop_delaymain}. In Section \ref{app:gradbound}, we provide gradient bounds for $\tau$-GRU. In Section \ref{app:uat}, we provide a universal approximation result for a general class of time-delayed RNNs. In Sections \ref{sect:appD}-\ref{app_tuningparam}, we provide additional experimental results and details. In Section \ref{sect_AppH}, we make some remarks on comparing our model with state-space models (SSMs).

We are using the following notation throughout this appendix: $\odot$ denotes Hadamard product, $| v |$ (or $\|v\|$) denotes vector norm for the vector $v$, $\| A \|$ denotes operator norm for the matrix $A$, $\| A \|_\infty$ denotes the matrix norm induced by $\infty$-norm (i.e., maximum absolute row sum of the matrix $A$), $\sigma$ and $\hat{\sigma}$ (or sigmoid) denote the $\tanh$ and sigmoid function, respectively, and $\lceil x \rceil$ and $\lfloor x \rfloor$ denote the ceiling function and floor function in $x$, respectively. 
In addition, $\delta_{i,j}$ denotes the Kronecker delta, i.e., $\delta_{i,j} = 1$ if $i=j$ and $\delta_{i,j} = 0$ if $i\neq j$.

\section{Illustrations of Differences Between ODE and DDE Dynamics}
\label{sect:appA}

Of particular interest to us are the differences between ODEs and DDEs that are driven by an input. 
To illustrate the differences in the context of RNNs in terms of how they map the input signal into an output signal, we consider the simple examples: 
\begin{itemize}
\item[(a)] DDE based RNNs with the hidden states $h \in \RR$ satisfying $\dot{h} = -h(t-\tau) + u(t)$, with $\tau = 0.5$ and $\tau = 1$, and $h(t) = 0$ for $t \in [-\tau,0]$, and
\item[(b)] an ODE based RNN with the hidden states $h \in \RR$ satisfying $\dot{h} = -h(t) + u(t)$, 
\end{itemize}
where $u(t) = \cos(t)$ is the driving input signal. 

\begin{figure}[!t]	
\centering	\includegraphics[width=0.85\textwidth]{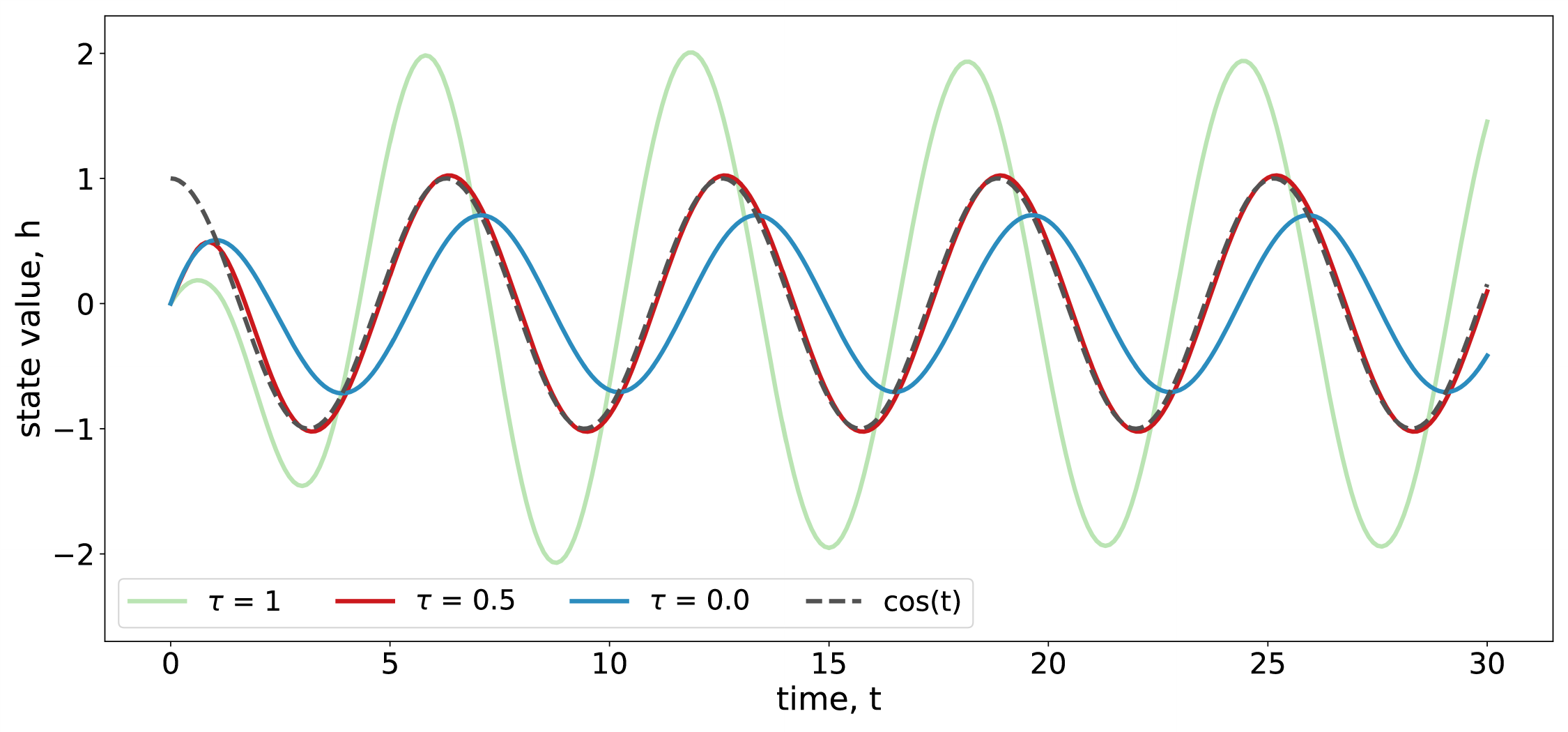}	\caption{Hidden state dynamics of the DDE based RNNs with $\tau=0.5$  and $\tau=1$, and the ODE based RNN ($\tau=0$). All RNNs are driven by the same cosine input signal. }	\label{fig:diff}
\end{figure}

Figure \ref{fig:diff} shows the difference between the dynamics of the hidden state driven by the input signal for (a) and (b). We see that, when compared to the ODE based RNN, the introduced delay causes time lagging in the response of the RNNs to the input signal. The responses are also amplified. In particular, using $\tau = 0.5$ makes the response of the RNN closely matches the input signal. In other words, simply fine tuning the delay parameter $\tau$ in the scalar RNN model is sufficient to replicate the dynamics of the input signal. 

To further illustrate the differences, we consider the following examples of RNN with a nonlinear activation: 
\begin{itemize}
\item[(c)] DDE based  RNNs with the hidden states $h \in \RR$ satisfying $\dot{h} = - h + \tanh(-h(t-\tau) + s(t))$, with $\tau > 0$, and $h(t) = 0$ for $t \in [-\tau,0]$, and 
\item[(d)] an ODE based  RNN with the hidden states $h \in \RR$ satisfying $\dot{h} = - h + \tanh(-h(t) + s(t))$, 
\end{itemize}
where the driving input signal $s$ is taken to be the truncated Weierstrass function:
\begin{equation}
    s(t) = \sum_{n=0}^3 a^{-n} \cos(b^n \cdot \omega t ),
\end{equation}
where $a=3$, $b=4$ and $\omega=2$.

\begin{figure}[!t]	
\centering	\includegraphics[width=0.85\textwidth]{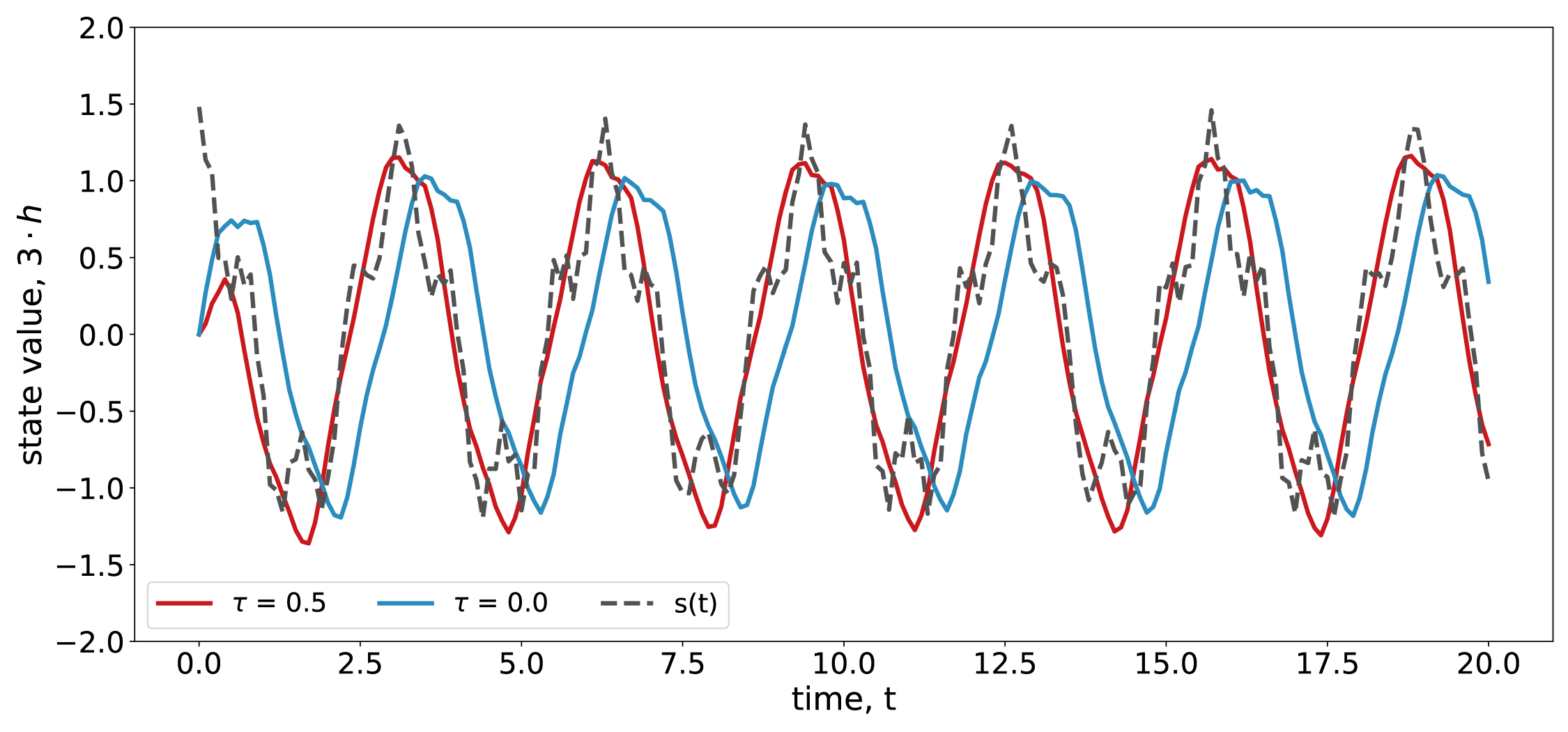}	\caption{Hidden state dynamics of the DDE based RNN with $\tau=0.5$ and the ODE based RNN ($\tau = 0$). All RNNs are driven by the same input signal $s(t)$. }	\label{fig:diff2}
\end{figure}

Figure \ref{fig:diff2} shows the difference between the input-driven dynamics of the  hidden states for (c) and (d). We see that, when compared to the ODE based RNN, the introduced delay causes time lagging  in the response of the RNNs to the input signal. Even though the response of both RNNs does not match the input signal precisely (since we consider RNNs with one-dimensional hidden states here and therefore their expressivity is limited), we see that using $\tau = 0.5$ produces a response that tracks the seasonality of the input signal better than the  ODE based RNN.

\clearpage
\section{Proof of Theorem \ref{thm_exist2main}} 
\label{sect:appB}

In this section, we provide a proof of Theorem \ref{thm_exist2main}. 
To start, note that one can view the solution of the DDE as a mapping of functions on the interval $[t -\tau,t]$ into functions on the interval $[t,t + \tau]$, i.e., as a sequence of functions defined over a set of contiguous time intervals of length $\tau$. 
This perspective makes it more straightforward to prove existence and uniqueness theorems analogous to those for ODEs than by directly viewing DDEs as an evolution over the state space $\RR^d$.

The following theorem, adapted from Theorem 3.7 in \citep{smith2011introduction}, provides sufficient conditions for existence and uniqueness of solution through a point $(t_0, \phi) \in \mathbb{R} \times C$ for the IVP \eqref{eq_genddemain}. Recall that  $C := C([-\tau, 0], \mathbb{R}^d)$, the Banach space of continuous functions from $[-\tau, 0]$ into $\mathbb{R}^d$ with the topology of uniform convergence. It is equipped with the norm $\|\phi \| := \sup \{ |\phi(\theta)| : \theta \in [-\tau, 0] \}$.

\begin{theorem}[Adapted from Theorem 3.7 in \citep{smith2011introduction}] 
\label{thm_existence}
Let $t_0 \in \RR$ and $\phi \in C$ be given. Assume that $f$ is continuous and satisfies the Lipschitz condition on each bounded subset of $\mathbb{R} \times C$, i.e., for all $a, b \in \mathbb{R}$, there exists a constant $K > 0$ such that 
\begin{equation}
    |f(t, \phi) - f(t, \psi)| \leq K \| \phi - \psi \|, \ t \in [a,b], \ \|\phi\|, \|\psi \| \leq M,
\end{equation}
with  $K$ possibly dependent on $a, b, M$. 

There exists $A > 0$, depending only on $M$ such that if $\phi \in C$ satisfies $\| \phi \| \leq M$, then there exists a unique solution $h(t) = h(t, \phi)$ of Eq. \eqref{eq_genddemain}, defined on $[t_0 - \tau, t_0 + A]$. Moreover, if $K$ is the Lipschitz constant for $f$ corresponding to $[t_0, t_0 + A]$ and M, then
\begin{equation}
    \max_{\eta \in [t_0 - \tau, t_0 + A]} |h(\eta, \phi) - h(\eta, \psi)| \leq \|\phi - \psi \| e^{KA}, \ \|\phi\|, \| \psi \| \leq M.
\end{equation}
\end{theorem}

We now provide existence and uniqueness result for the continuous-time $\tau$-GRU model, assuming that the input $x$ is continuous in $t$. As before, we define the state $h_t \in C$ as:
\begin{equation}
    h_t(\theta) := h(t+\theta), \ -\tau \leq \theta \leq 0.
\end{equation}
Then the DDE defining the model can be formulated as the following IVP for the nonautonomous system: 
\begin{equation} 
    \dot{h}(t) = -h(t) + u(t, h(t)) + a(t, h(t)) \odot z(t, h_t), \ t \geq t_0,
\end{equation}
and $h_{t_0} = \phi \in C$ for some initial time $t_0 \in \mathbb{R}$, with the dependence on $x(t)$ casted as dependence on $t$ for notational convenience.

Now we restate Theorem \ref{thm_exist2main} from the main text and provide the proof.

\begin{theorem}[Existence and uniqueness of solution for continuous-time $\tau$-GRU] \label{thm_exist2}
Let $t_0 \in \RR$ and $\phi \in C$ be given. There exists a unique solution $h(t) = h(t, \phi)$ of Eq. \eqref{eq_gendde}, defined on $[t_0 - \tau, t_0 + A]$ for any $A > 0$. In particular, the solution exists for all $t \geq t_0$, and
\begin{equation}
    \| h_t(\phi) - h_t(\psi) \| \leq \| \phi - \psi \| e^{K(t-t_0)},
\end{equation}
for all $t \geq t_0$, where $K = 1 + \|W_1\| +  \|W_2\| + \|W_4\|/4$.
\end{theorem}
\begin{proof}
We shall apply Theorem \ref{thm_existence}. 
To verify the Lipschitz condition: for any $\phi, \psi \in C$,
\begin{align}
    &
    \hspace{-7mm}
    |(u(t, \phi) + a(t, \phi) \odot z(t, \phi) - \phi) - (u(t, \psi) + a(t, \psi) \odot z(t, \psi) - \psi)| \nonumber \\
    &\leq |u(t, \phi) - u(t, \psi)| + |\phi - \psi| + |(a(t,\phi) - a(t, \psi)) \odot z(t,\phi) | + | a(t, \psi) \odot (z(t, \phi) - z(t, \psi) )| \\
    &\leq \|W_1\| \cdot |\phi - \psi| + |\phi - \psi| + \frac{1}{4} \|W_4\| \cdot  |\phi - \psi| + \|W_2\| \cdot  |\phi - \psi| \\
    &=: K |\phi - \psi|,
\end{align}
where we have used the  fact that the tanh and sigmoid are Lipschitz continuous, both bounded by one in absolute value, and they have  positive derivatives of magnitude no larger than one and $1/4$ respectively  in the last inequality~above.

Therefore, we see that the right hand side function satisfies a global Lipschitz condition and so the result follows from Theorem \ref{thm_existence}. 
\end{proof}

\section{Proof of Proposition \ref{prop_delaymain}} \label{sect:appC}

In this section, we restate Proposition \ref{prop_delaymain}, and  provide its proof and some remarks.

\begin{proposition} \label{prop_delay}
Consider the linear time-delayed RNN  whose hidden states are described by the update equation:
\begin{equation} \label{eq_recursion}
    h_{n+1} = Ah_n + Bh_{n-m} + Cu_n, \ n=0,1,\dots,
\end{equation}
and $h_n = 0$ for $n =-m, -m+1, \dots, 0$ with $m > 0$. 
Then, assuming that $A$ and $B$ commute, we have:
\begin{equation}
    \frac{\partial h_{n+1}}{\partial u_i} = A^{n-i} C, 
\end{equation}
for $n=0,1, \dots, m$, $i=0,\dots, n$, and 
\begin{align}
    \frac{\partial h_{m+1+j}}{\partial u_i} &= A^{m+j-i} C + \delta_{i,j-1} BC + 2 \delta_{i,j-2} ABC \nonumber \\
    &\ \ \ \ + 3 \delta_{i,j-3} A^2 B C + \cdots + j \delta_{i,0} A^{j-1} B C, 
\end{align}
for $j = 1,2,\dots, m+1$, $i=0,1,\dots, m+j$, where $\delta_{i,j}$ denotes the Kronecker delta.
\end{proposition}

\begin{proof}
Note that by definition $h_i = 0$ for $i=-m, -m+1, \dots, 0$, and, upon iterating the recursion \eqref{eq_recursion}, one~obtains:
\begin{equation}
    h_{n+1} = A^n C u_0 + A^{n-1} C u_1 + \cdots + AC u_{n-1} + C u_n,
\end{equation}
for $n = 0, 1, \dots, m$.

Now, applying the above formula for $h_1$, we obtain 
\begin{align}
h_{m+2} &= A h_{m+1} + B h_1 + Cu_{m+1} \nonumber \\
&= (B + A^{m+1} ) C u_0 + A^{m} C u_1 + \cdots + A^2 C u_{m-1} + ACu_m + Cu_{m+1}.
\end{align}

Likewise, we obtain:
\begin{align}
    h_{m+3} &= A h_{m+2} + B h_2 + Cu_{m+2} \nonumber \\
    &= (BA + A^{m+2} + AB) C u_0 + (B+A^{m+1})C u_1 + A^m C u_2 + \cdots + AC u_{m+1} + C u_{m+2} \nonumber \\
    &= (2AB + A^{m+2}) C u_0 + (B+A^{m+1})C u_1 + A^m C u_2 + \cdots + AC u_{m+1} + C u_{m+2},
\end{align}
where we have used commutativity of $A$ and $B$ in the last line above.

Applying the above procedure repeatedly and using commutativity of $A$ and $B$ give:
\begin{equation} \label{eq_linear}
    h_{m+1+j} = (A^{m+j} + j A^{j-1} B) C u_0 + ( A^{m+j-1} + (j-1) A^{j-2} B) C u_1 + \cdots + ACu_{m+j-1} + C u_{m+j}, 
\end{equation}
for $j = 1,2, \dots, m+1$.

The formula in the proposition then follows upon taking the derivative with respect to the $u_i$ in the above formula for the hidden states $h_k$.
\end{proof}

We remark that one could also derive formula for the gradients $\frac{\partial h_{n+1+j}}{\partial u_i}$ for $n \geq 2m+1$ and those for our $\tau$-GRU architecture analogously, albeit the resulting expression is quite complicated.
In particular, the dependence on higher powers of $B$ for the coefficients in front of the Kronecker deltas would appear in the formula for the former case (with much more complicated expressions without assuming commutativity of the matrices). However, we emphasize that the qualitative conclusion derived from the analysis remains the same: the introduction of delays places more emphasis on gradient information due to input elements in the past, so they act as buffers to propagate the gradient information more effectively than the counterpart models without delays.

\section{Gradient Bounds for \texorpdfstring{$\tau$}{tau}-GRU}
\label{app:gradbound}

\noindent 
{\bf On the exploding and vanishing gradient problem.} 
For simplicity of our discussion here, we consider the loss~function: 
\begin{equation}
    \mathcal{E}_n = \frac{1}{2} \| y_n - \overline{y}_n  \|^2,
\end{equation}
where $n = 1, \dots, N$ and $\overline{y}_n$ denotes the underlying growth truth. The training of $\tau$-GRU involves computing gradients of this loss function with respect to its underlying parameters $\theta \in \Theta = [W_{1,2,3,4}, U_{1,2,3,4}, V]$ at each iteration of the gradient descent algorithm. Using chain rule, we obtain \citep{pascanu2013difficulty}:
\begin{equation}
    \frac{\partial \mathcal{E}_n}{\partial \theta} = \sum_{k=1}^n \frac{\partial \mathcal{E}_n^{(k)}}{\partial \theta},
\end{equation}
where 
\begin{equation}
    \frac{\partial \mathcal{E}_n^{(k)}}{\partial \theta} = \frac{\partial \mathcal{E}_n}{\partial h_n} \frac{\partial h_n}{\partial h_k} \frac{\partial^+ h_k}{\partial \theta},
\end{equation}
with $\frac{\partial^+ h_k}{\partial \theta}$ denoting  the ``immediate'' partial derivative of
the state $h_k$ with respect to $\theta$, i.e., where $h_{k-1}$ is taken as a constant with respect to $\theta$ \citep{pascanu2013difficulty}.

The partial gradient $\frac{\partial \mathcal{E}_n^{(k)}}{\partial \theta}$  measures the contribution to the hidden state gradient at step $n$ due to the information at step $k$. It can be shown that this gradient behaves as \begin{equation}
\frac{\partial \mathcal{E}_n^{(k)}}{\partial \theta} \sim \gamma^{n-k},
\end{equation}
for some $\gamma > 0$ \citep{pascanu2013difficulty}. If $\gamma > 1$, then the gradient grows exponentially in sequence length, for long-term dependencies where $k \ll n$, causing the exploding gradient problem. On the other hand, if $\gamma < 1$, then the gradient decays exponentially for $k \ll n$, causing the vanishing gradient problem.  Therefore,  we can investigate how $\tau$-GRU deals with these problems by deriving bounds on the gradients. In particular, we are interested in the behavior of the gradients for long-term dependencies, i.e., $k \ll n$, and shall show that the delay mechanism in $\tau$-GRU slows down the exponential decay rate, thereby reducing the sensivity to the vanishing gradient problem.

Recall that the update equations defining $\tau$-GRU are given by $h_n = 0$ for $n=-m, -m+1, \dots, 0$, 
\begin{equation}
    h_n = (1-g(A_{n-1})) \odot h_{n-1} + g(A_{n-1}) \odot [u(B_{n-1}) + a(C_{n-1}) \odot z(D_{n-m-1})],
\end{equation}
for $n=1,2,\dots, N$, where $m := \lfloor \tau/\Delta t \rfloor \in \{1,2,\dots, N-1\}$, $A_{n-1} = W_3 h_{n-1} + U_3 x_{n-1}$, $B_{n-1} = W_1 h_{n-1} + U_1 x_{n-1}$, 
$C_{n-1} = W_4 h_{n-1} + U_4 x_{n-1}$, and $D_{n-m-1} = W_2 h_{n-m-1} + U_2 x_{n-1}$.

In the sequel, we shall denote the $i$th component of a vector $v$ as $v^i$ and the $(i,j)$ entry of a matrix $A$ as $A^{ij}$.

We start with the following lemma.

\begin{lemma} \label{app_lem}
For every $i$, we have $ h_n^i = 0$, for $n = -m, -m+1, \dots, 0$, and $ |h_n^i| \leq 2$, for $n=1,2,\dots, N$.
\end{lemma}

\begin{proof}
The $i$th component of the hidden states of $\tau$-GRU are given by: $h_n^i = 0$ for $n=-m, -m+1, \dots, 0$, and
\begin{equation}
    h_n^i = (1-g(A^i_{n-1}))  h^i_{n-1} + g(A^i_{n-1})  [u(B^i_{n-1}) + a(C^i_{n-1}) z(D^i_{n-m-1})],
\end{equation}
for $n=1,2,\dots, N$.

Using the fact that $g(x), a(x) \in (0,1)$ and $u(x), z(x) \in (-1,1)$ for all $x$, we can bound the $h_n^i$ as:
\begin{align}
    h_n^i &\leq (1-g(A^i_{n-1}))  \max(h^i_{n-1},2) + g(A^i_{n-1}) \max(h^i_{n-1},2) \nonumber \\
    &\leq \max(h^i_{n-1},2),
\end{align}
for all $i$ and $n = 1,2,\dots, N$.

Similarly, we have:
\begin{align}
    h_n^i &\geq (1-g(A^i_{n-1}))  \min(-2, h^i_{n-1}) + g(A^i_{n-1}) \min(-2, h^i_{n-1}) \nonumber \\
    &\geq \min(-2, h^i_{n-1}),
\end{align}
for all $i$ and $n = 1,2,\dots, N$.

Thus, 
\begin{equation}
      \min(-2, h^i_{n-1}) \leq h_n^i \leq \max(h^i_{n-1},2),
\end{equation}
for all $i$ and $n = 1,2,\dots, N$.

Now, iterating over $n$ and using $h_0^i = 0$ for all $i$, we obtain $-2 \leq h_n^i \leq 2$ for all $i$ and $n = 1,2,\dots, N$.

\end{proof}

We now provide the gradients bound for $\tau$-GRU in the following proposition and proof.

\begin{proposition} \label{app_prop}
Assume that there exists an $\epsilon > 0$ such that $\max_n g(A^i_{n-1}) \geq \epsilon$ and   $\max_n a(C^i_{n-1}) \geq \epsilon$ for all $i$. Then
\begin{align}
    \left \| \frac{\partial h_n}{\partial h_k} \right\|_\infty &\leq (1+C-\epsilon)^{n-k} \nonumber \\ &\ \ \ \ \ + \| W_2\|_\infty \cdot \left( (1+C-\epsilon)^{n-k-2} \delta_{m,1} + \dots + (1+C-\epsilon) \delta_{m, n-k-2} + \delta_{m, n-k-1} \right),
\end{align}
for $n=1, \dots, N$ and $k < n$, where $C = \|W_1\|_\infty +  \|W_3\|_\infty + \frac{1}{4}\|W_4\|_\infty$.
\end{proposition}
\begin{proof}
Recall $h_n = 0$ for $n=-m, -m+1, \dots, 0$, and
\begin{equation}
    h_n = (1-g(A_{n-1})) \odot h_{n-1} + g(A_{n-1}) \odot [u(B_{n-1}) + a(C_{n-1}) \odot z(D_{n-m-1})] =: F(h_{n-1}, h_{n-m-1}),
\end{equation}
for $n=1,2,\dots, N$.

Denote $q_{n,l} := \frac{\partial F}{\partial h_{n-l}}$, where $F := F(h_{n-1}, h_{n-l})$ for $l>1$. 
The gradients $\frac{\partial h_n}{\partial h_k}$ can be computed recursively as~follows.

\begin{align}
    p_n^{(1)} &:= \frac{\partial h_n}{\partial h_{n-1}},  \\
    p_n^{(2)} &:= \frac{\partial h_n}{\partial h_{n-2}} = p_n^{(1)} p_{n-1}^{(1)} + q_{n,2} \delta_{m,1}, \\  p_n^{(3)} &:= \frac{\partial h_n}{\partial h_{n-3}} = p_n^{(1)} p_{n-1}^{(2)} + q_{n,3} \delta_{m,2}, \\
    \vdots \\
    p_n^{(n-k)} &:= \frac{\partial h_n}{\partial h_{k}} = p_n^{(1)} p_{n-1}^{(n-k-1)} + q_{n,n-k} \delta_{m,n-k-1}.
\end{align}
As $\|p_n^{(n-k)}\| \leq \|p_n^{(1)} \| \cdot \| p_{n-1}^{(n-k-1)}\| + \|q_{n,n-k}\| \delta_{m,n-k-1}$, it remains to upper bound the $p_n^{(1)}$ and $q_{n,n-k}$.

The $i$th component of the hidden states can be written as:
\begin{equation}
    h_n^i = (1-g(A^i_{n-1}))  h^i_{n-1} + g(A^i_{n-1})  [u(B^i_{n-1}) + a(C^i_{n-1})  z(D^i_{n-m-1})],
\end{equation}
where $A^i_{n-1} = W_3^{iq} h_{n-1}^q + U_3^{ir} x_{n-1}^{r}$, $B^i_{n-1} = W_1^{iq} h_{n-1}^q + U_1^{ir} x_{n-1}^{r}$, $C^i_{n-1} = W_4^{iq} h_{n-1}^q + U_4^{ir} x_{n-1}^{r}$, and $D^i_{n-m-1} = W_2^{iq} h_{n-m-1}^q + U_2^{ir} x_{n-1}^{r}$, using Einstein's summation notation for repeated indices.

Therefore, applying chain rule and using Einstein's summation for repeated indices in the following, we obtain:
\begin{align}
 \frac{\partial h_n^i}{\partial h_{n-1}^j} &= (1-g(A_{n-1}^i))  \frac{\partial h_{n-1}^i}{\partial h_{n-1}^j} - \frac{\partial g(A_{n-1}^i)}{\partial A_{n-1}^l} \frac{\partial A_{n-1}^l}{\partial h_{n-1}^j} h_{n-1}^i + g(A_{n-1}^i) \frac{\partial u(B_{n-1}^i)}{\partial B_{n-1}^l} \frac{\partial B_{n-1}^l}{\partial h_{n-1}^j} \nonumber \\
 &\ \ \ \ \ + \frac{\partial g(A_{n-1}^i)}{\partial A_{n-1}^l} \frac{\partial A_{n-1}^l}{\partial h_{n-1}^j} u(B_{n-1}^i) +    g(A_{n-1}^i) \frac{\partial a(C_{n-1}^i)}{\partial C_{n-1}^l} \frac{\partial C_{n-1}^l}{\partial h_{n-1}^j} z(D_{n-m-1}^i) \nonumber \\
 &\ \ \ \ \ + \frac{\partial g(A_{n-1}^i)}{\partial A_{n-1}^l} \frac{\partial A_{n-1}^l}{\partial h_{n-1}^j} a(C_{n-1}^i) z(D_{n-m-1}^i).
\end{align}

Noting that $\frac{\partial g(A_{n-1}^i)}{\partial A_{n-1}^l} = 0$, $\frac{\partial u(B_{n-1}^i)}{\partial B_{n-1}^l} = 0$ and $\frac{\partial a(C_{n-1}^i)}{\partial C_{n-1}^l} = 0$ for $i \neq l$, we have:
\begin{align}
 \frac{\partial h_n^i}{\partial h_{n-1}^j} &= (1-g(A_{n-1}^i))  \frac{\partial h_{n-1}^i}{\partial h_{n-1}^j} - \frac{\partial g(A_{n-1}^i)}{\partial A_{n-1}^i} \frac{\partial A_{n-1}^i}{\partial h_{n-1}^j} h_{n-1}^i + g(A_{n-1}^i) \frac{\partial u(B_{n-1}^i)}{\partial B_{n-1}^i} \frac{\partial B_{n-1}^i}{\partial h_{n-1}^j} \nonumber \\
 &\ \ \ \ \ + \frac{\partial g(A_{n-1}^i)}{\partial A_{n-1}^i} \frac{\partial A_{n-1}^i}{\partial h_{n-1}^j} u(B_{n-1}^i) +    g(A_{n-1}^i) \frac{\partial a(C_{n-1}^i)}{\partial C_{n-1}^i} \frac{\partial C_{n-1}^i}{\partial h_{n-1}^j} z(D_{n-m-1}^i) \nonumber \\
 &\ \ \ \ \ + \frac{\partial g(A_{n-1}^i)}{\partial A_{n-1}^i} \frac{\partial A_{n-1}^i}{\partial h_{n-1}^j} a(C_{n-1}^i) z(D_{n-m-1}^i) \\
 &= (1-g(A_{n-1}^i)) \delta_{i,j}  - \frac{\partial g(A_{n-1}^i)}{\partial A_{n-1}^i} W_3^{ij} h_{n-1}^i + g(A_{n-1}^i) \frac{\partial u(B_{n-1}^i)}{\partial B_{n-1}^i} W_1^{ij} \nonumber \\
 &\ \ \ \ \ + \frac{\partial g(A_{n-1}^i)}{\partial A_{n-1}^i} W_3^{ij} u(B_{n-1}^i) +    g(A_{n-1}^i) \frac{\partial a(C_{n-1}^i)}{\partial C_{n-1}^i} W_4^{ij}  z(D_{n-m-1}^i) \nonumber \\
 &\ \ \ \ \ + \frac{\partial g(A_{n-1}^i)}{\partial A_{n-1}^i} W_3^{ij} a(C_{n-1}^i) z(D_{n-m-1}^i).
\end{align}

Using the assumption that $\max_n g(A^i_{n-1}),  \max_n a(C^i_{n-1}) \geq \epsilon$ for all $i$, the fact that $|z(x)|, |u(x)| \leq 1$, $g(x), a(x) \in (0,1)$, $g'(x), a'(x) \leq 1/4$, $u'(x) \leq 1$ for all $x \in \RR$, and Lemma \ref{app_lem},  we obtain:
\begin{align}
 \left|\frac{\partial h_n^i}{\partial h_{n-1}^j} \right| &\leq (1-\epsilon) \delta_{i,j} + \frac{1}{4} |W_3^{ij}| \cdot |h_{n-1}^i| + |W_1^{ij}| +  \frac{1}{4} |W_3^{ij}| + \frac{1}{4} |W_4^{ij}| + \frac{1}{4} |W_3^{ij}| \\
 &\leq (1-\epsilon) \delta_{i,j} + |W_1^{ij}| + |W_3^{ij}| + \frac{1}{4} |W_4^{ij}|.
\end{align}

Therefore, 
\begin{equation}
     \left \| \frac{\partial h_n}{\partial h_{n-1}} \right\|_\infty := \max_{i=1,\dots, d} \sum_{j=1}^d \left| \frac{\partial h_n^i}{\partial h_{n-1}^j } \right| \leq (1-\epsilon ) +  \|W_1\|_\infty + \|W_3 \|_\infty + \frac{1}{4} \|W_4\|_\infty =: 1-\epsilon + C.
\end{equation}

Likewise, we obtain, for $l>1$:
\begin{align}
    \frac{\partial F^i}{\partial h_{n-l}^j} &= g(A_{n-1}^i) a(C_{n-1}^i) \frac{\partial z(D_{n-l}^i)}{\partial D_{n-l}^i} \frac{\partial D_{n-l}^i}{\partial h_{n-l}^j}  \\
    &= g(A_{n-1}^i) a(C_{n-1}^i) \frac{\partial z(D_{n-l}^i)}{\partial D_{n-l}^i} W_2^{ij}.
\end{align}

Using the fact that $|g(x)|, |a(x)| \leq 1$ and $|z'(x)| \leq 1$ for all $x \in \RR$, we obtain:
\begin{align}
    \left|\frac{\partial F^i}{\partial h_{n-l}^j}\right| &\leq    |W_2^{ij}|,
\end{align}
for $l > 1$,
and thus
$\|q_{n,n-k}\|_\infty = \| \frac{\partial F}{\partial h_k} \|_\infty \leq \|W_2\|_\infty$ for $k > 1$.

The upper bound in the proposition follows by using the above bounds for $\|p_{n}^{(1)}\|_\infty$ and $\|q_{n,n-k}\|_\infty$,  and iterating the recursion $\|p_n^{(n-k)}\| \leq \|p_n^{(1)} \| \cdot \| p_{n-1}^{(n-k-1)}\| + \|q_{n,n-k}\| \delta_{m,n-k-1}$ over $k$.

\end{proof}

From Proposition \ref{app_prop}, we see that if $\epsilon > C$, then the gradient norm decays exponentially as $k$ becomes large. However,  the delay in $\tau$-GRU introduces jump-ahead connections  (buffers) to slow down the exponential decay. For instance, choosing $m=1$ for the delay, we have  $\left \| \frac{\partial h_n}{\partial h_k} \right\|  \sim (1+C-\epsilon)^{n-k-2}$ as $k \to \infty$ (instead of $\left \| \frac{\partial h_n}{\partial h_k} \right\|  \sim (1+C-\epsilon)^{n-k-1}$ as $k \to \infty$ in the case when no delay is introduced into the model). The larger the $m$ is, the more effective the delay is able to slow down the exponential decay of the gradient norm. These qualitative conclusions can already be derived by studying the linear time-delayed RNN, which we consider in the main text  for simplicity.

\section{Approximation Capability of Time-Delayed RNNs} \label{app:uat}

RNNs (without delay) have been shown to be universal approximators of a large class of open dynamical systems \citep{schafer2006recurrent}. Analogously, RNNs with delay can be shown to be universal approximators of open dynamical systems with~delay. 

Let $m > 0$ (time lag) and consider the state space models (which, in this section, we shall simply refer to as delayed RNNs) of the form:
\begin{align} \label{eq_gendRNN}
    s_{n+1} &= f(As_n + B s_{n-m} + Cu_n + b), \nonumber \\
    r_n &= Ds_n,
\end{align}
and dynamical systems of the form 
\begin{align} \label{ds_approx}
x_{n+1} &= g(x_n, x_{n-m}, u_n), \nonumber \\
o_n &= o(x_n),
\end{align}
for $n = 0,1,\dots, N$.
Here $u_n \in \RR^{d_u}$ is the input, $o_n \in \RR^{d_o}$ is the target output of the dynamical systems to be learned, $s_n \in \RR^d$ is the hidden state of the learning model, $r_n \in \RR^{q}$ is the model output, $f$ is the tanh function applied component-wise, the maps $g$ and $o$ are Lipschitz continuous, and the matrices $A$, $B$, $C$, $D$ and the vector $b$ are learnable parameters. For simplicity, we take the initial functions to be $s_n = y_n = 0$ for $n = -m, -m+1, \dots, 0$.

The following theorem shows that  the delayed RNNs  \eqref{eq_gendRNN} are capable of approximating a large class of time-delay dynamical systems, of the form \eqref{ds_approx}, to arbitrary accuracy.

\begin{theorem} \label{thm_uat}
Assume that there exists a constant $R > 0$ such that $\max(\|x_{n+1}\|, \|u_n\|) < R$ for $n = 0,1,\dots, N$. Then, for a given $\epsilon > 0$, there exists a delayed RNN of the form  \eqref{eq_gendRNN} such that the following holds for some $d$:
\begin{equation} \label{thm_bd}
    \|r_n - o_n \| \leq \epsilon,  
\end{equation}
for $n = 0,1,\dots, N$.
\end{theorem}

\begin{proof}
The proof proceeds along the line of \citep{schafer2006recurrent, rusch2022long}, using the universal approximation theorem (UAT) for feedforward neural network maps and  with straightforward modification to deal with the extra delay variables $s_{n-n}$ and $x_{n-m}$ here.   The proof  proceeds in a similar manner as the one provided in Section E.4 in \citep{rusch2022long}.
The goal is to construct hidden states, output state, weight matrices and bias vectors such that an output of the delayed RNN approximates the dynamical system \eqref{ds_approx}.

Let $\epsilon > \epsilon^* > 0, R^* > R  \gg 1$ be parameters to be defined later. Then, using the UAT for continuous functions with neural networks with the tanh activation function \citep{barron1993universal}, we can obtain the following statements.  Given an $\epsilon^*$, there exist weight matrices $W_1$, $W_2$, $W_3$, $V_1$ and a bias vector $b_1$ of appropriate dimensions such that the neural network defined by $\mathcal{N}_1(h, \tilde{h}, u) := W_3 \tanh(W_1 h + W_2 \tilde{h} + V_1 u + b_1)$ approximates the underlying function $g$ as follows:
\begin{equation}
    \max_{\max (\|h\|, \|\tilde{h} \|, \|u\|) < R^* } \| g(h, \tilde{h}, u)  - \mathcal{N}_1(h, \tilde{h}, u) \| \leq \epsilon^*.
\end{equation}

Now, we define the dynamical system:
\begin{align}
    p_{n} &= W_3 \tanh(W_1 p_{n-1}   + W_2 p_{n-m-1} + V_1 u_{n-1} + b_1),
\end{align}
with $p_i = 0$ for $i = -m, -m+1, \dots, 0$.

Then using the above approximation bound, we obtain, for $n=1, \dots, N+1$,
\begin{align}
    \|x_{n} - p_{n} \| &= \| g(x_{n-1}, x_{n-m-1}, u_{n-1})  - p_n \| \\
    &\leq \|g(x_{n-1}, x_{n-m-1}, u_{n-1}) - W_3 \tanh(W_1 p_{n-1}   + W_2 p_{n-m-1} + V_1 u_{n-1} + b_1)\| \\
    &\leq \|g(x_{n-1}, x_{n-m-1}, u_{n-1}) - g(p_{n-1}, p_{n-m-1},  u_{n-1}) \| \nonumber \\
    &\ \ \ \ \ + \|  g(p_{n-1}, p_{n-m-1},  u_{n-1}) - W_3 \tanh(W_1 p_{n-1}   + W_2 p_{n-m-1} + V_1 u_{n-1} + b_1)\| \\
    &\leq Lip(g) (\| x_{n-1} - p_{n-1} \| + \|x_{n-m-1} - p_{n-m-1}\|) +  \epsilon^*,
\end{align}
where $Lip(g)$ is the Lipschitz constant of $g$ on the compact set $\{ (h,\tilde{h}, u): \| h\|, \|\tilde{h}\|, \|u\| < R^* \}$.

Iterating the above inequality over $n$ leads to:
\begin{equation}
    \|x_n - p_n \|\leq \epsilon^* C_1(n, m, Lip(g)),
\end{equation}
for some constant $C_1>0$ that is dependent on $n, m, Lip(g)$.

Using the Lipschitz continuity of the output function $o$, we obtain: 
\begin{equation} \label{out_b}
    \| o_n - o(p_n) \| \leq \epsilon^*C_2(n, m, Lip(g), Lip(o)),
\end{equation}
for some constant $C_2$ that is dependent on $n, m, Lip(g), Lip(o)$, where $Lip(o)$ is the Lipschitz constant of $o$ on the compact set $\{ h : \| h\|  < R^* \}$.

Next we use the UAT for neural networks again to obtain the following approximation result. Given an $\overline{\epsilon}$, there exists weight matrices $W_4, W_5$ and bias vector $b_2$ of appropriate dimensions such that the tanh neural network, $\mathcal{N}_2(h) := W_5 \tanh(W_4 h + b_2)$ approximates the underlying output function $o$ as:
\begin{equation}
    \max_{\|h\| < R^*} \| o(h) - \mathcal{N}_2(h) \| \leq \overline{\epsilon}.
\end{equation}

Defining $\overline{o}_n = W_5 \tanh(W_4 p_n + b_2)$, we obtain, using the above approximation bound and the inequality \eqref{out_b}:
\begin{equation} \label{o_b_2}
    \|o_n - \overline{o}_n \| = \|o_n - o(p_n) \| + \| o(p_n) - \overline{o}_n \|    \leq \epsilon^* C_2(n, m, Lip(g), Lip(o)) + \overline{\epsilon}.
\end{equation}

Now, let us denote: 
\begin{align}
    \tilde{p}_{n} &= \tanh(W_1 p_{n-1}   + W_2 p_{n-m-1} + V_1 u_{n-1} + b_1),
\end{align}
so that $p_n = W_3 \tilde{p}_n$. With this notation, we have:
\begin{equation}
\overline{o}_n = W_5 \tanh(W_4 W_3 \tanh(W_1 W_3 \tilde{p}_{n-1}   + W_2 W_3 \tilde{p}_{n-m-1} + V_1 u_{n-1} + b_1)  + b_2).
\end{equation}

Since the function $R(y) = W_5 \tanh(W_4 W_3 \tanh(W_1 W_3 y + W_2 W_3 \tilde{p}_{n-m-1} + V_1 u_{n-1} + b_1) + b_2)$ is Lipschitz continuous in $y$, we can apply the UAT again to obtain: for any $\tilde{\epsilon}$, there exists weight matrices $W_6, W_7$ and bias vector $b_3$ of appropriate dimensions such that 
\begin{equation}
    \max_{\|y\| < R^*} \| R(y) - W_7 \tanh(W_6 y + b_3) \| \leq \tilde{\epsilon}.
\end{equation}
Denoting $\tilde{o}_n := W_7 \tanh(W_6 p_{n-1} + b_3)$ and using the above approximation bound, we obtain $\|\overline{o}_n - \tilde{o}_n \| \leq \tilde{\epsilon}$.

Finally, we collect all the ingredients above to construct a delayed RNN that can approximate the dynamical system \eqref{ds_approx}.
To this end, we define the hidden states (in an enlarged state space):  $s_n := (\tilde{p}_n, \hat{p}_n)$, with $\tilde{p}_n$, $\hat{p}_n$ sharing the same dimension.
These hidden states evolve according to the dynamical system:
\begin{align}
s_n &= \tanh\left( 
  \left[ {\begin{array}{cc}
    W_1 W_3 & 0 \\
    W_6 W_3 & 0 \\
  \end{array} } \right] s_{n-1} +  \left[ {\begin{array}{cc}
    W_2 W_3 & 0 \\
    0 & 0 \\
  \end{array} } \right] s_{n-m-1} +  \left[ {\begin{array}{c}
    V_1 u_{n-1}  \\
    0 \\
  \end{array} } \right] +  \left[ {\begin{array}{c}
    b_1 \\
    0 \\
  \end{array} } \right] 
\right).
\end{align}
Defining the output state as $r_n := [0, W_7] s_n$, with the $s_n$ satisfying the above system, we arrive at a delayed RNN that approximates the dynamical system  \eqref{ds_approx}. In fact, we can  verify that $r_n = \tilde{o}_n$. 
Setting $\overline{\epsilon} < \epsilon/2$ and  $\epsilon^* < \epsilon/(2 C_2(n, m, Lip(g), Lip(o)))$ give us the  bound \eqref{thm_bd} in the theorem.
\end{proof}

\section{Additional Details and Experiments}
\label{sect:appD}

In this section, we provide additional empirical results and details to demonstrate the advantages of $\tau$-GRU when compared to other RNN architectures. As with many RNN papers, we report the performance for the single best model in our experiments. Rather than running all the baseline models ourselves, we report the results from the corresponding papers since we assume that the authors have done the best job tuning their respective model. In a few cases we train the baseline models and indicate the results by a “*”.

\subsection{Copy Task}

Here we consider the copy task as an additional benchmark problem to assess the ability of $\tau$-GRU to handle long-range dependencies.
The copy task was originally proposed by~\cite{hochreiter1997long}, and requires the model to memorize the initial $10$ elements of the input sequence for the duration of $T$ time steps. Then the model is tasked with outputting the initial elements as accuratelty as possible.

\begin{figure}[!ht]
	\centering
	\includegraphics[width=0.84\textwidth]{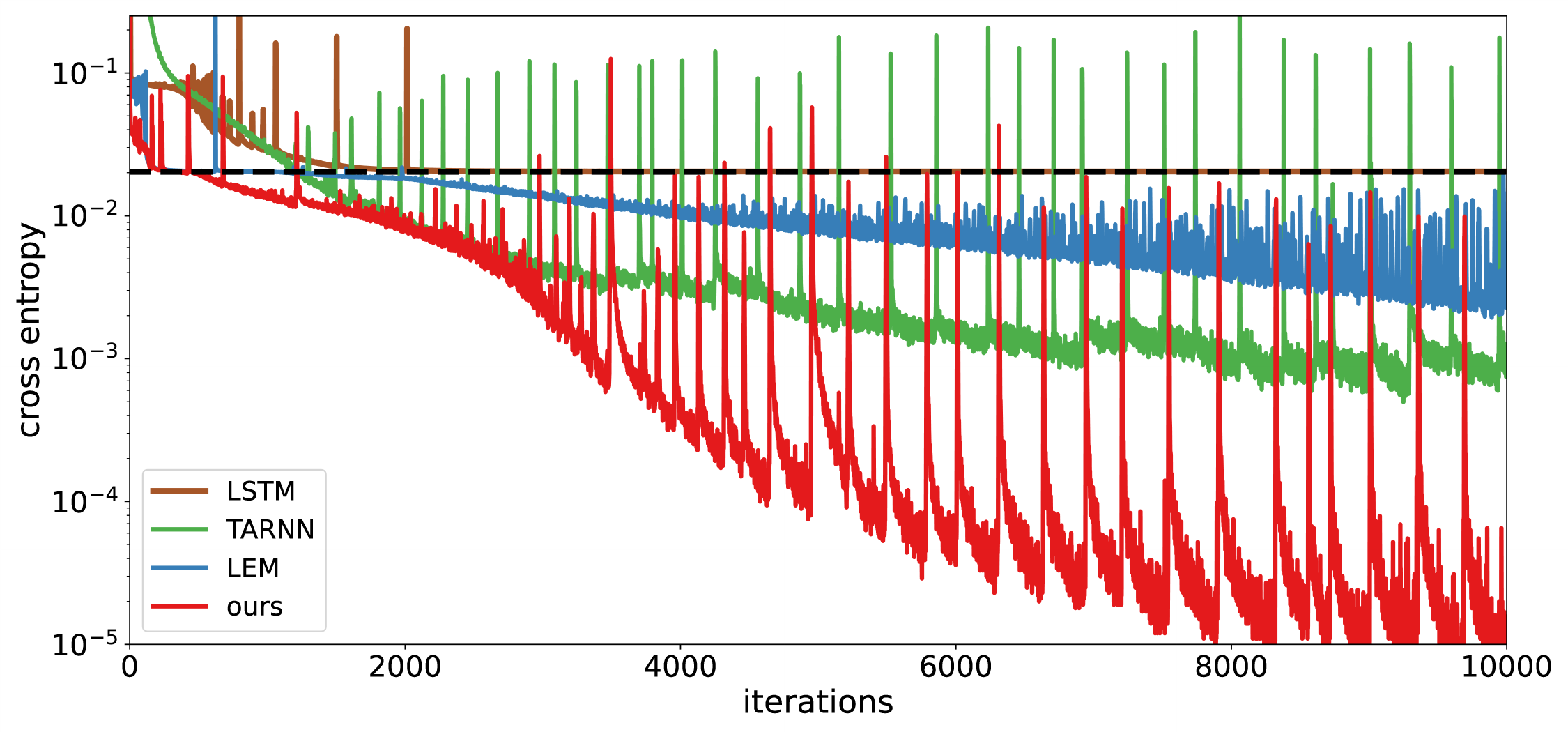}
	\caption{Cross-entropy as a function of iterations for the copy task for different recurrent models. Here we consider the copy task with a sequence lengths $T=1000$.}
	\label{fig:copy}
\end{figure}

Here we consider a long sequence with $T=1000$. The results are shown in Figure~\ref{fig:copy}. It can be seen that our $\tau$-GRU is able to solve the problem, whereas other models such as LEM or TARNN show inferior performance.

\subsection{Speech Recognition: Google 12}

Here, we consider the Google Speech Commands data set V2~\citep{warden2018speech} to demonstrate the performance of our model for speech recognition. 
The aim of this task is to learn a model that can classify a short audio sequence, which is sampled at a rate of 16 kHz from 1 second utterances of $2,618$ speakers. 
We consider the Google 12-label task (Google12) which is composed of 10 keyword classes, and in addition one class that corresponds to `silence', and a class corresponding to `unknown' keywords. 
We adopt the standard train/validation/test set split for evaluating our model, and we use dropout, applied to the inputs, with rate 0.03 to reduce overfitting.

Table~\ref{tab:results_google12} presents the results for our $\tau$-GRU and a number of competitive RNN architectures. 
We adopt the results for the competitive RNNs from~\citep{rusch2022long}. Our proposed $\tau$-GRU shows the best performance on this task, i.e., $\tau$-GRU is able to outperform gated and continuous-time RNNs on this task that requires an expressive recurrent unit.

\begin{table}[!ht]
	\caption{Test accuracy results for Google12. Results indicated by $^*$ are produces by us, results indicated by $^+$ are from~\citep{rusch2022long}. }
	\label{tab:results_google12}
	\centering
	\scalebox{0.9}{
		\begin{tabular}{l c  c ccccccc}
			\toprule
			Model & Accuracy ($\%$) &  \# units & \# param \\			
			\midrule 
			tanh RNN~\citep{rusch2022long}$^+$ & 73.4 & 128 & 27k \\
			LSTM~\citep{rusch2022long}$^+$  & 94.9 & 128 & {107k}\\
			GRU~\citep{rusch2022long}$^+$  & 95.2 & 128 & 80k \\			
			AsymRNN~\citep{chang2018antisymmetricrnn}$^+$ & 90.2 & 128 & 20k \\
			expRNN~\cite{lezcano2019cheap}$^+$ & 92.3 & 128 & 19k \\
			coRNN~\citep{rusch2021coupled}$^+$ & 94.7 & 128 & 44k \\
			Fast GRNN~\citep{kusupati2018fastgrnn}$^+$ & 94.8 & 128 & 27k \\
			LEM~\citep{rusch2022long} & 95.7 & 128 & {107k} \\
			Lipschitz RNN~\citep{erichson2020lipschitz}$^*$ & 95.6 & 128 & 34k\\
			Noisy RNN~\citep{lim2021noisy}$^*$ & 95.7 & 128 & 34k\\
			iRNN~\citep{Kag2020RNNs}$^*$ & 95.1 & - & {8.5k} \\ 
			TARNN~\citep{kag2021time}$^*$ & 95.9 & 128 & {107k} \\ 
			\midrule
			\textbf{ours} & \textbf{96.2} & 128 & {107k} \\
			\bottomrule
	\end{tabular}}
        \vspace{-0.2cm}
\end{table}

\subsection{Learning the Dynamics of Mackey-Glass System} \label{app_mackey}

Here, we consider the task of learning the Mackey-Glass equation, originally introduced in \citep{mackey1977oscillation} to  model the variation in the relative quantity of mature cells in the blood:
\begin{equation}
    \dot{x} = a \frac{x(t-\delta)}{1 + x^n(t-\delta)} - b x(t), \ t \geq \delta,
\end{equation}
where $\delta \geq 17$, $a, b, n > 0$, with $x$ satisfying $\dot{x} = a x(0)/(1 + x(0)^n) - b x$ for $t \in [0, \delta]$. It is a scalar equation with chaotic dynamics, with infinite-dimensional state space. Increasing the value of $\delta$ increases the dimension of the~attractor.

\begin{figure}[!ht]
	\centering
	\includegraphics[width=0.44\textwidth]{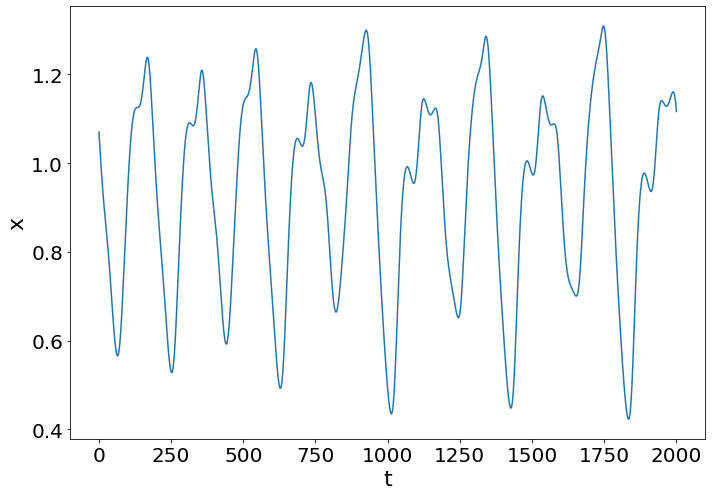}
	\includegraphics[width=0.44\textwidth]{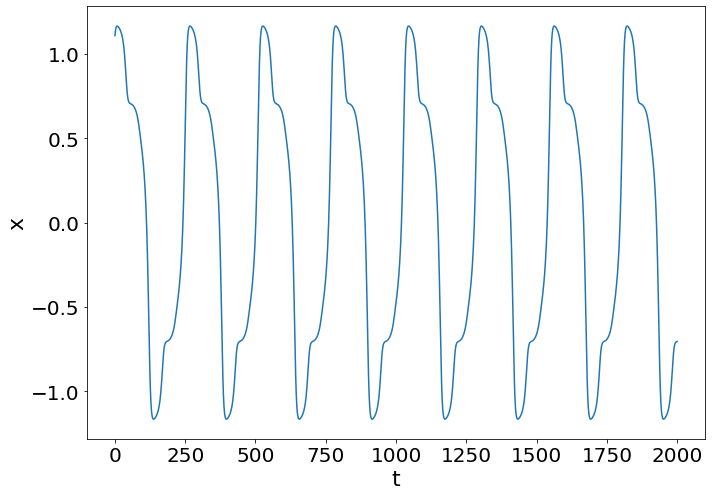}
	\caption{A realization of the Mackey-Glass dynamics (left) and the DDE based ENSO dynamics (right). }
	\label{fig:data_plot}
\end{figure}

For data generation, we choose $a = 0.2$, $b = 0.1$, $n=10$, $\delta = 17$, $x(0) \sim \rm{Unif}(0,1)$, and use the classical Runge-Kutta  method (RK4) to integrate the system numerically from $t=0$ to $t = 1000$ with a step-size of 0.25. The training and testing samples are the time series (of length 2000) generated by the RK4 scheme on the interval $[500,1000]$ for different realizations of $x(0)$. Figure \ref{fig:data_plot} shows a realization of the trajectory produced by the Mackey-Glass system (and also the DDE based ENSO system considered in the main text).

Table \ref{tab:results_MG} shows that our $\tau$-GRU model (with $\alpha = \beta = 1$ and using $\tau = 10$) is more effective in learning the Mackey-Glass system when compared to other RNN architectures. We also see that the predictive performance deteriorates without making full use of the combination of the standard recurrent unit and delay recurrent unit  (setting either $\alpha$ or $\beta$ to zero). Moreover, $\tau$-GRU demonstrates improved performance when compared to the simple delay GRU model (Eq. \eqref{eq_simpleDRNN}) and the counterpart model without using the gating. Similar observation also holds for the ENSO prediction task; see Table \ref{tab:add_results_ENSO}.

\begin{table}[!t] 
	\caption{Results for the Mackey-Glass system prediction.}
	\label{tab:results_MG}
	\centering
	\scalebox{0.9}{
		\begin{tabular}{l c  c  ccccccc}
			\toprule
			Model & MSE ($\times 10^{-2}$) &  \# units & \# parameters \\			
			\midrule 
			Vanilla RNN & 0.3903 & 16 & 0.321k \\
			LSTM & 0.6679 & 16 & 1.233k \\
			GRU & 0.4351 & 16 & 0.929k \\
			Lipschitz RNN & 8.9718 & 16 & 0.561k \\
			coRNN & 1.6835 & 16 & 0.561k \\
			LEM & 0.1430 & 16 & 1.233k \\
			\midrule
			simple delay  GRU (Eq. \eqref{eq_simpleDRNN})  & 0.2772 & 16 & 0.897k \\
			ablation (no gating) & 0.2765 & 16 & 0.929k \\
			ablation ($\alpha = 0$) & 0.1553 & 16 & 0.625k \\
			ablation ($\beta = 0$) & 0.2976 & 16 & 0.929k \\
			\midrule
			\textbf{$\tau$-GRU (ours)} & \textbf{0.1358}  & 16 & 1.233k \\
			\bottomrule
	\end{tabular}}
\end{table}

\begin{figure}[!ht]
	\centering
	\includegraphics[width=0.44\textwidth]{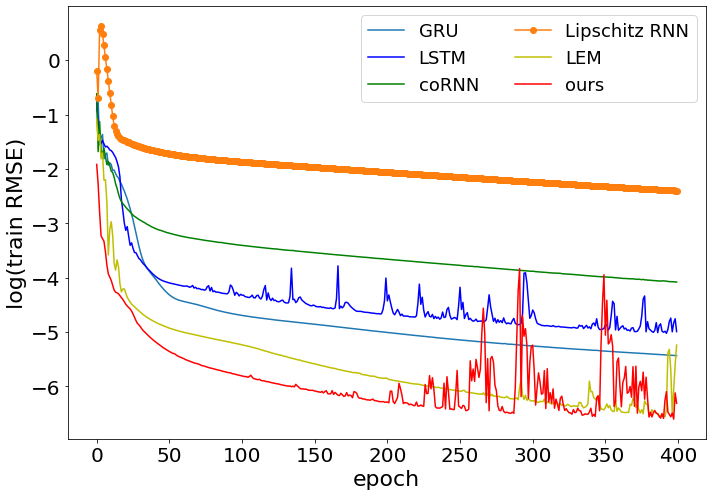}
	\includegraphics[width=0.44\textwidth]{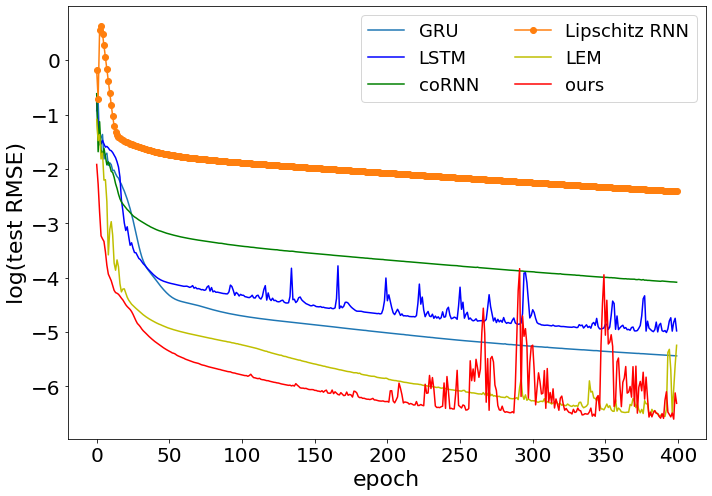}
	\caption{Train RMSE (left) and test RMSE (right) vs. epoch for the Mackey-Glass learning task.}
	\label{fig:mg_traintest}
\end{figure}

\begin{figure}[!ht]
	\centering
	\includegraphics[width=0.44\textwidth]{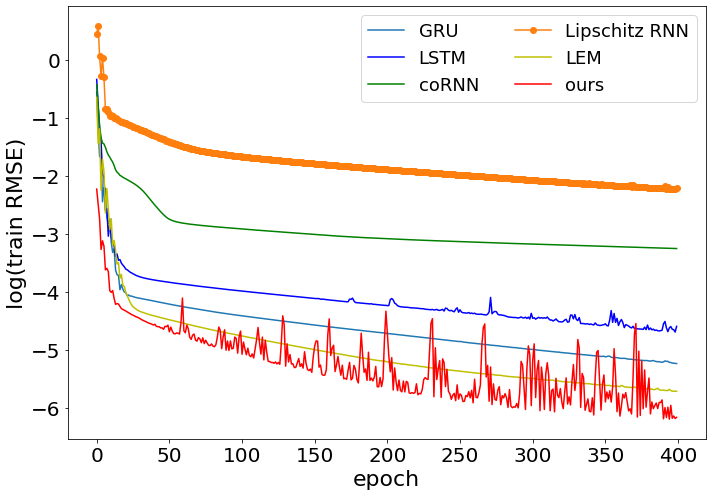}
	\includegraphics[width=0.44\textwidth]{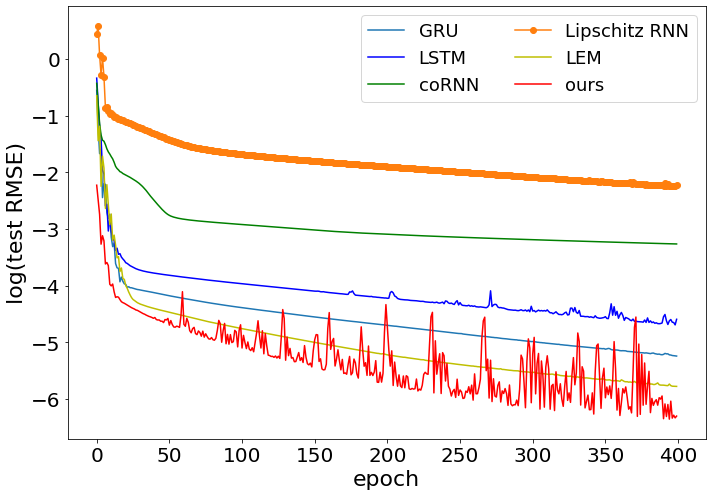}
	\caption{Train RMSE (left) and test RMSE (right) vs. epoch for the ENSO  learning task.}
	\label{fig:mz_traintest}
\end{figure}

Figure \ref{fig:mg_traintest} shows that our model converges much faster than other RNN models during training. In particular, our model is able to achieve both lower training and testing error (as measured by the root mean square error (RMSE)) with fewer epochs, demonstrating the effectiveness of the delay mechanism in improving the performance on the problem of long-term dependencies. This is consistent with our analysis on how the gradient
information is propagated through the delay buffers in the network (see Proposition \ref{prop_delaymain}), suggesting that the delay buffers can propagate gradient more efficiently. Similar behavior is also observed for the ENSO task; see Figure \ref{fig:mz_traintest}.

\begin{table}[!ht]
	\caption{Additional results for the ENSO model prediction.}
	\label{tab:add_results_ENSO}
	\centering
	\scalebox{0.9}{
		\begin{tabular}{l c  c  ccccccc}
			\toprule
			Model & MSE ($\times 10^{-2}$) &  \# units & \# parameters \\			
			\midrule 
			
		    simple delay GRU (Eq. \eqref{eq_simpleDRNN}) & 0.2317 & 16 & 0.897k \\
			ablation (no gating) & 0.4289 & 16 & 0.929k \\
		\midrule
		\textbf{$\tau$-GRU (ours)} & \textbf{0.17} & 16 & 1.2k \\
    	\bottomrule
	\end{tabular}}
\end{table}

\section{Tuning Parameters} \label{app_tuningparam}

To tune our $\tau$-GRU, we use a non-exhaustive random search within the following plausible ranges for $\tau={5,\dots,200}$. We used Adam as our optimization algorithm for all of the experiments. For the synthetic data sets generated by the ENSO and Mackey-Glass system, we used learning rate of 0.01. For the other experiments we considered learning rates between 0.01 and 0.0005. We used dropout for the IMDB and Google12 task to avoid overfitting.

Table~\ref{tab:tuning} is listing the tuning parameters for the different tasks that we considered in this work.

\begin{table}[!ht]
	\caption{Summary of tuning parameters.}
	\label{tab:tuning}
	\centering
	\scalebox{0.85}{
		\begin{tabular}{l c c c c c c c c }
			\toprule
			Name           &  d & lr  & $\tau$ & dropout & epochs \\
      		\midrule 
			Adding Task $N=2000$ & 128 & 0.0026 & 900 & - & 200 \\
            Adding Task $N=5000$ & 128 & 0.002 & 2000 & - & 200 \\
   			\midrule 
			IMDB  & 128 & 00012 & 1 & 0.04 & 30 \\
			\midrule 
			HAR-2  & 128 & 0.00153 & 10 & - & 100 \\
			\midrule    
			sMNIST  & 128 & 0.0018 & 50 & - & 60 \\
			\midrule 
			psMNIST & 128 & 0.0055 & 65 & - & 80 \\		
			\midrule
			sCIFAR & 128 & 0.0035 & 30 & - & 50  \\
            \midrule 
			nCIFAR & 128 & 0.0022 & 965 & - & 50 \\
			\midrule
   			Google12 & 128 & 0.00089 & 5 & 0.03  &  60 \\
            \midrule
            ENSO & 16  & 0.01  & 20 & - & 400  \\
            \midrule
            Mackey-Glass system  & 16 & 0.01  & 10 & - &  400  \\
   \bottomrule
	\end{tabular}}
\end{table}

\paragraph{Sensitivity to Random Initialization.}

We evaluate our models for each tasks using 8 seeds. The maximum, minimum, average values, and standard deviations obtained for each task are tabulated in Table~\ref{tab:minmax}.

\begin{table}[h]
	\caption{Sensitivity to random initialization evaluated over 8 runs with different seeds.}
	\label{tab:minmax}
	\centering
	\scalebox{0.91}{
		\begin{tabular}{l c c c c c c c c}
			\toprule
			  Task     & Maximum & Minimum  & Average & standard dev. & standard error & d \\
			\midrule 
            IMDB  & 88.7	& 86.2	 & 87.9 & 0.82 & 0.29 & 128 \\
            HAR-2  & 97.4	& 96.6 & 96.9 &	0.41 & 0.17 & 128 \\
			sMNIST  & 99.4 & 99.1 &	99.3 & 0.08 & 0.02 & 128 \\
            psMNIST  & 97.3 & 96.0 & 96.8 &	0.39 & 0.13 & 128 \\
            sCIFAR  & 74.9 & 72.65 & 73.54 & 0.90 & 0.41 & 128 \\
            nCIFAR  & 62.7 & 61.7 &	62.3 & 0.32 & 0.11 & 128 \\
            Google12 & 96.2 & 95.7 & 95.9 &	0.17 & 0.05 & 128 \\
			\bottomrule
	\end{tabular}}
\end{table}

\section{\texorpdfstring{$\tau$}{tau}-GRU is More Parameter-Efficient than SSMs in the Small Data Regime} \label{sect_AppH}

The state-space models (SSMs) are a class of models which uses structured linear RNNs representations as layers in a deep learning pipeline \citep{gu2021combining, gu2021efficiently, voelker2019legendre}. They originate from  the HIPPO framework which offers an optimal solution to a natural online function approximation problem  \citep{gu2020hippo}.  While they have been shown to demonstrate state-of-the-art sequence modeling performance in several domains, these models may not be optimal for tasks in the small data regime, which is quite common in many scientific applications where the available data is often scarce. Therefore, developing  lightweight yet effective models that can perform well in this scenario is  valuable. 

To support this claim, we conduct an additional experiment to
compare the performance of S4 model \citep{gu2021efficiently} and LMU \citep{voelker2019legendre} with our proposed model in the small data regime. We consider the Mackey-Glass system prediction problem in Section \ref{app_mackey}, where the size of training set is 128 and the length of each sequence sample is 2000. Table \ref{table_com} shows that $\tau$-GRU significantly outperforms the other two models while being parameter efficient for this task despite having to deal with a relatively small training set consisting of long sequences.

\begin{table}[!ht] 
\caption{Comparison of the performance, in terms of test mean squared error (MSE), of $\tau$-GRU with S4 and LMU on the Mackey-Glass system prediction problem.}
\vspace{0.3cm}
\centering
\label{table_com}
\begin{tabular}{l c c}
\toprule
Model & MSE & \# Parameters \\
\midrule
S4 \citep{gu2021efficiently}   & 0.0097 & 1.2k \\
LMU \citep{voelker2019legendre} & 0.0064 & 1.2k \\
{\bf $\tau$-GRU} & {\bf 0.0014} & {\bf 1.2k} \\
\bottomrule
\end{tabular}
\end{table}

\noindent {\bf Discussion.}
The above results show that there  are domains/settings where nonlinear RNNs (i.e., RNNs using a nonlinear activation function in the hidden state update equation) are beneficial, while there are other domains/settings where SSMs are favorable. For instance, SSMs can outperform $\tau$-GRU in sequential image classification, albeit using higher number of trainable parameters \citep{gu2021efficiently}. Also, SSMs have the tendency to require more data for training to perform well as compared to nonlinear RNNs. SSMs need to be stacked to model non-linear dynamics, and thus they are typically deep \citep{wang2023state}. 
A fair comparison between various forms of SSMs \citep{gu2023mamba, hasani2022liquid, smith2022simplified, orvieto2023resurrecting} and the RNNs we consider is challenging, because those SSMs are deep whereas the RNNs we consider are shallow.

\end{document}